\documentclass[]{fairmeta}

\usepackage{microtype}
\usepackage{graphicx}
\usepackage{subcaption}
\usepackage{booktabs} %

\usepackage{hyperref}

\usepackage{tcolorbox}

\usepackage{amsmath}
\usepackage{amssymb}
\usepackage{mathtools}
\usepackage{amsthm}

\definecolor{myred}{HTML}{CB4154}

\usepackage{amsmath}
\usepackage{bbm}
\usepackage{booktabs}
\newcommand{\indicator}[1]{\mathbbm{1}\left\lbrace #1\right\rbrace}
\newcommand{\thrsoracle}{\tau_{\text{oracle}}}
\newcommand{\thrsapprox}{\tau_{\text{approx}}}
\newcommand{\thrsstatic}{\tau_{\text{static}}}

\newcommand{\tbulk}{T^{\text{bulk}}}

\newcommand{\oraclecriterion}{c^h_{\text{oracle}}}

\usepackage{listings}
\usepackage{xcolor}
\definecolor{keyword}{rgb}{0.75, 0.13, 0.13}
\definecolor{comment}{rgb}{0.25, 0.5, 0.35}
\definecolor{string}{rgb}{0.6, 0.1, 0.1}
\usepackage{listings}
\usepackage{xcolor}

\definecolor{codeblue}{rgb}{0.13, 0.13, 0.75}
\definecolor{codegray}{rgb}{0.5, 0.5, 0.5}
\definecolor{codepurple}{rgb}{0.58, 0.0, 0.82}
\definecolor{backcolour}{rgb}{0.98, 0.98, 0.98}

\definecolor{functionblue}{RGB}{67,110,238}    %
\definecolor{variablegreen}{RGB}{102,153,0}    %
\definecolor{torchfunc}{RGB}{255,140,0}        %
\definecolor{outputpurple}{RGB}{147,112,219}   %
\definecolor{keywordcolor}{RGB}{0,119,170}     %

\definecolor{lightgreen}{RGB}{76,175,80}     %
\definecolor{darkgreen}{RGB}{46,125,50}      %
\definecolor{torchfunc}{RGB}{255,140,0}      %
\definecolor{outputpurple}{RGB}{147,112,219} %

\usepackage{listings}
\usepackage{xcolor}

\definecolor{backcolour}{rgb}{0.99,0.99,0.99}
\definecolor{codegray}{rgb}{0.5,0.5,0.5}
\definecolor{codeblue}{rgb}{0.0,0.0,0.5}
\definecolor{codepurple}{rgb}{0.58,0.0,0.82}
\definecolor{orange}{rgb}{1.0,0.5,0.0}

\theoremstyle{plain}

\theoremstyle{definition}

\theoremstyle{remark}

\usepackage[textsize=tiny]{todonotes}

\title{Unveiling Simplicities of Attention:\\ Adaptive Long-Context Head Identification}

\author[2,*]{Konstantin Donhauser}
\author[1]{Charles Arnal}
\author[1]{Mohammad Pezeshki}
\author[1]{Vivien Cabannes}
\author[1]{David Lopez-Paz}
\author[1]{Kartik Ahuja}

\affiliation[1]{FAIR at Meta}
\affiliation[2]{ETH Zurich}

\contribution[*]{Work done at Meta}

\abstract{
The ability to process long contexts is crucial for many natural language processing tasks, yet it remains a significant challenge. While substantial progress has been made in enhancing the efficiency of attention mechanisms, there is still a gap in understanding how attention heads function in long-context settings. In this paper, we observe that while certain heads consistently attend to local information only, others swing between attending to local and long-context information depending on the query. This raises the question: can we identify which heads require long-context information to predict the next token accurately? We demonstrate that it's possible to predict which heads are crucial for long-context processing using only local keys. The core idea here is to exploit a simple model for the long-context scores via second moment approximations. These findings unveil simple properties of attention in the context of long sequences, and open the door to potentially significant gains in efficiency.
}
\usepackage{amsthm}
\usepackage{amsmath}
\usepackage{amssymb}
\usepackage{algorithm}
\usepackage{algorithmic}
\usepackage{subcaption}

\date{\today}
\correspondence{First Author at \email{konstantin.donhauser@ai.ethz.ch}}

\renewcommand{\paragraph}[1]{\textbf{#1}}
\usepackage{amsmath}

\begin{document}

\maketitle

\section{Introduction}
\label{section:intro}

The landscape of large language models (LLMs) is rapidly evolving, with modern architectures capable of generating text from vast contexts. Recent advances have led to a significant increase in context window sizes, with Llama 3 \citep{dubey2024llama}, DeepSeekv3 \citep{liu2024deepseek}, and Gemini \citep{team2024gemini} supporting windows of at least 128k. 
However, long context modeling still poses significant challenges \citep{hsieh2024ruler} in terms of both accuracy  and  the substantial cost of processing long contexts in terms of memory and run-time compute. 
In spite of their importance, our current comprehension of the attention mechanism in long-context tasks remains incomplete. This work aims to address some of these knowledge gaps.

\begin{figure*}[ht]
\centering
\begin{subfigure}[b]{\linewidth}
\centering
        \includegraphics[width=1.0\linewidth]{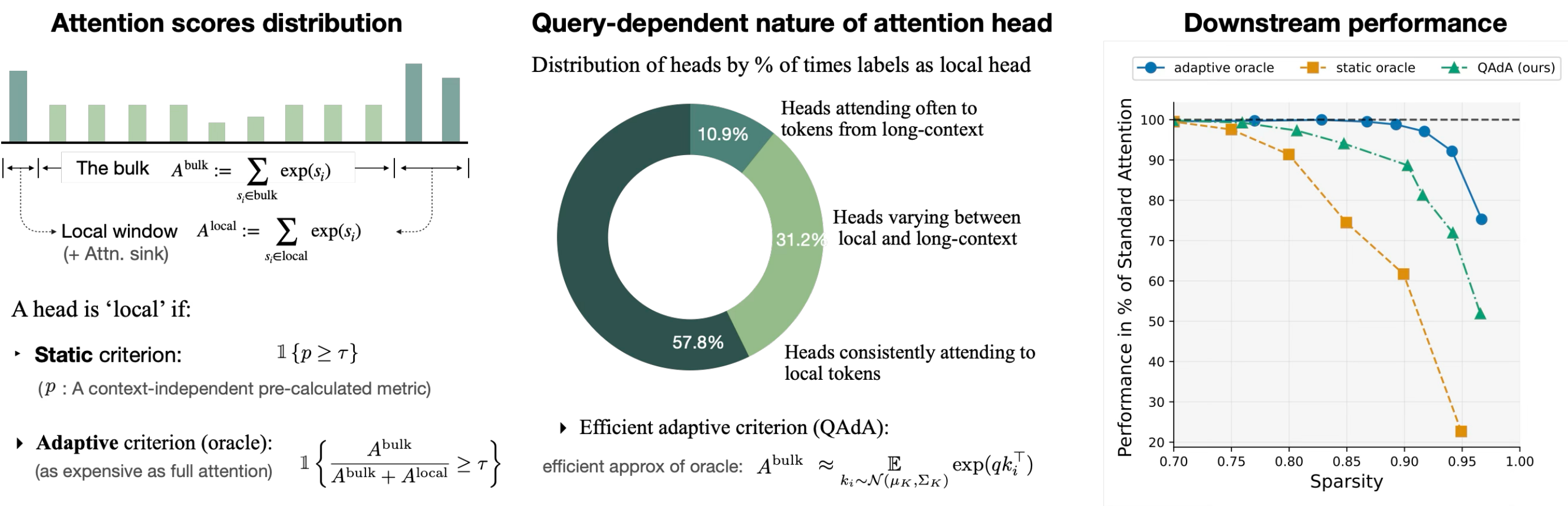}\\
        \begin{subfigure}[b]{\linewidth}
    \end{subfigure}
\end{subfigure}
    \vspace{-1cm}
     \caption{
    \small{ 
    \textbf{Attention sparsity and its impact on efficiency.} 
    \textit{Left:} Attention scores are split into \textit{bulk} ($A^{\text{bulk}}$) for distant tokens and \textit{local window} ($A^{\text{local}}$) for nearby ones. A head is considered local if most of its attention mass falls within the local window. The static criterion pre-assigns local heads, while the adaptive oracle query-dependently compares bulk and local contributions but is computationally expensive. Our approximation models $A^{\text{bulk}}$ using a Gaussian distribution for efficiency.
    \textit{Middle:} Oracle-based classification with $\tau = 0.6$ (see Figure~\ref{fig:compare-approx} for the threshold) reveals three types of heads: consistently local (heads labeled more than $95\%$ of the times as local), often long-context (less than $50\%$), and varying, which switch behavior dynamically.
    \textit{Right:} Comparison of three methods: Static (green) removes a fixed fraction of heads, the adaptive oracle (blue) dynamically selects heads but is costly, and our adaptive method (purple) achieves near-oracle performance with significantly lower cost. As sparsity increases, static pruning degrades performance, while our adaptive method remains robust.
    These results show that \textit{most attention heads do not need to attend to the entire context}, enabling significant efficiency gains with \textit{query-adaptive} head classification.} }
    \label{fig:gaussian}
\end{figure*}

Despite the overwhelming complexity of state-of-the-art models, certain simple behaviors in the attention mechanism are strikingly consistent. In particular, many forms of sparse behaviors have been consistently observed, and exploited by numerous methods for efficient inference (see Section~\ref{sec:relatedworks}). 
Among them, \citet{xiao2023efficient} showed that even when computing the attention only using tokens close to the current token plus initial ``sink'' tokens, as illustrated in Figure~\ref{fig:gaussian},  the model is still capable of generating fluent text. We refer to these tokens as local window, and always implicitly include the   initial tokens as they play a crucial role as an attention ``sink'' (see also \citet{chen2024magicpig,gu2024attention,sun2024massive}). 

However, such a local window approximation, if applied to every attention head simultaneously, necessarily harms the capabilities of LLMs to retrieve and process long-context information (see e.g., \citet{xiao2024duoattention}).
Instead, to overcome such limitations, we aim to identify the heads whose output can be well-approximated via a local window attention, and apply the approximation to those only. If a head can be approximated via a local approximation, we call it a \textbf{local head}, and otherwise it is a \textbf{long-context head}. 
In particular, we ask:     Which heads can be approximated using a local window with minimal impact on downstream task performance?

Two approaches to this problem can be distinguished:
  \textit{Static} criteria label the heads -- local vs long-context --  once for all queries, while \textit{query-adaptive} criteria change the labels from query to query.  Static criteria, as used by \citet{xiao2024duoattention,tang2024razorattention}, have the advantage that all key-value pairs (except for the few in the local window) of local heads can be discarded, thus saving memory. While recent works \citep{wu2024retrieval,tang2024razorattention,hong2024token} 
 provide some evidence that a \textit{fixed} small subset of the heads are particularly relevant for processing long-context information, the following question remains unclear:
 \begin{center}
     \textit{How much sparsity (measured as the average percentage of local heads) can we gain using query-adaptive criteria compared to static criteria?}
     \end{center}

\paragraph{Contribution 1.} We present an extensive analysis comparing a query-adaptive oracle criterion, which selects local heads independently for each token, with static criteria. We make two observations: first, we find that static criteria can label up to 60\% of the heads as local heads without impacting downstream task evaluations, which confirms the intuition from \citep{wu2024retrieval}. Nevertheless, we find that a query-adaptive oracle criterion allows to label a substantially higher percentage of heads as local heads (up to 90\%) without sacrificing performance (see Figure~\ref{fig:gaussian}).

Unfortunately, the oracle requires the computation of the full attention scores. This leads to the following question:
\begin{center}
    \textit{ For each query, can we determine which heads are long-context and which are local without computing the full attention scores?}
\end{center}

The relevance of this question is twofold: on one hand, answering it helps guide further research toward developing more compute-efficient attention mechanisms. On the other hand, it advances our understanding of the inner workings of attention mechanisms, which is central to mechanistic interpretability (see, e.g., \citet{olsson2022context}). 

\paragraph{Contribution 2.} We address this question by proposing a novel query-adaptive attention criterion (QAdA) based on second-order statistics of the attention scores (briefly summarized in Figure~\ref{fig:gaussian}).
Our experiments on three families of LLMs, Llama \citep{dubey2024llama}, Qwen \citep{bai2023qwen} and Mistral \citep{jiang2023mistral} applied to a variety of standard long-context benchmarks, as well as hard  reasoning tasks embedded in long-context prompts, show that this relatively simple criterion allows to efficiently identify long-context heads: our method increased sparsity at a smaller loss in downstream performance than oracle static approaches. 
Along with our other experiments, it sheds light onto certain simple behaviors of attention heads in long-context settings.

\begin{figure*}
    \centering
    \includegraphics[width=\textwidth]{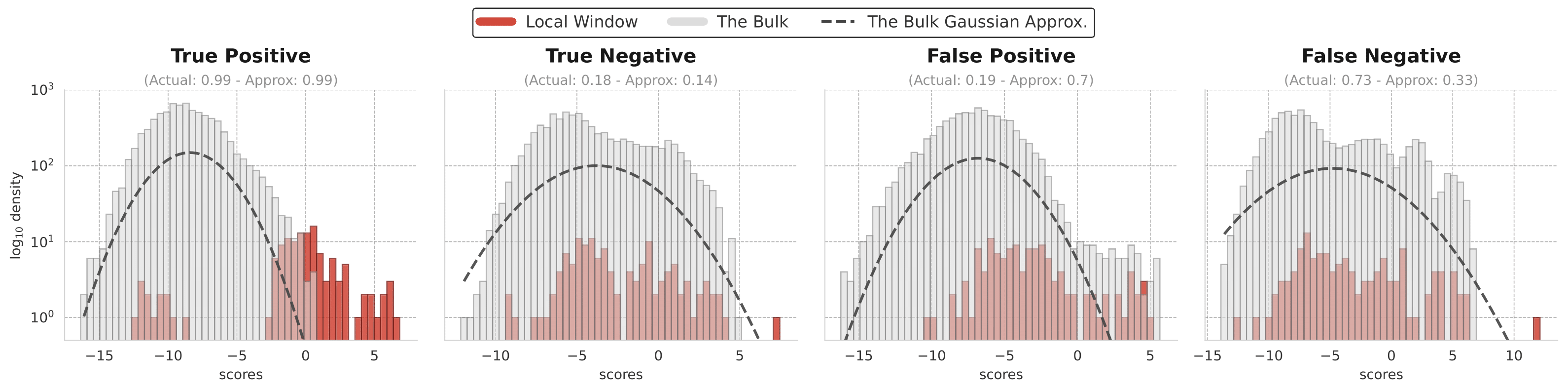}
    \caption{
        Examples of attention score distributions for each possible outcome with 
        $\thrsapprox = \thrsoracle = 0.6$ with the oracle criterion as ground truth. 
        We show histograms of scores from the \textbf{local window} $\mathcal{I}$ (\textcolor{brown}{brown}) 
        and the \textbf{bulk complement} $[T] \setminus \mathcal{I}$ (\textcolor{gray}{gray}), 
        along with the bulk Gaussian approximation (black dashed line). 
        The annotations above each plot indicate the values taken by the statistics used for the oracle criterion 
        and the adaptive criterion.
    }
    \label{fig:scores}
\end{figure*}

\section{Preliminaries}
\label{sec:setting}
 We consider decoder-only transformer models \citep{vaswani2017attention},  consisting of $L$-layers each containing one attention  and one feed-forward block, using the rotary positional encoding (RoPE, \citet{su2024roformer}),  which is commonly used in state-of-the-art open source LLMs, e.g., Llama3 \citep{dubey2024llama}, Qwen \citep{bai2023qwen} or Gemma \citep{team2024gemma}. 
During inference, when predicting the next token, every single attention head  takes as input a vector of (already rotated) queries $q\in \mathbb R^{1\times d}$  and the (updated and rotated) cached key-value pairs $K, V \in \mathbb R^{T\times d}$, with sequence length $T$, and returns the weighted average of the values:
\begin{equation}
\label{eq:softmax}
    o = \text{softmax}(s) V \quad \text{with scores}\quad s= q K^\top/ \sqrt{d}
\end{equation}

\paragraph{Local window approximation.} 
We are interested in long-context settings, where $T$ is large. For a given query and attention head, one can restrict the head's attention to a \textit{local window}: instead of computing the head's attention scores with respect to each of the $T$ keys, only the attention scores corresponding to the first $T_{\text{sink}}$ input tokens (i.e. those closest to the start of the sequence) and the last $T_{\text{local}} - T_{\text{sink}}$ tokens are computed (as illustrated in Figure~\ref{fig:gaussian}) and used to produce the output, where $T_{\text{local}}, T_{\text{sink}} \in \mathbb{N}$ are fixed parameters. Though they may not contain particularly relevant information, the first $ T_{\text{sink}}$ tokens are included to serve as ``attention sink'' tokens, in line with the observations from \citet{xiao2023efficient}.
To summarize it more formally, we call $\mathcal I := \{1,\ldots, T_{\text{sink}}\} \cup \{T -  T_{\text{local}} + T_{\text{sink}}+1, \ldots, T\} \subset [T] $ the set of local indices, and the output of an attention head restricted to a local window is equal to $o_{\text{local}} = \text{softmax}(s_{\mathcal I}) V_{\mathcal I}, $ with $s_{\mathcal I} = q K_{\mathcal I}^\top /\sqrt{d}$.

 \paragraph{Query-adaptive oracle criterion} 
 To determine which heads are local, we need to define a criterion that makes a decision for each query. 
 We call the heads labeled by the criterion  \textbf{local head} (for a given input token) and the others \textbf{long-context head}.
Assuming that we have access to all scores, a natural way to define such a criterion is to compare the mass of attention scores from the local window $\mathcal I$ to some threshold. That is, given a threshold $\thrsoracle$, an attention head $h$, and its associated attention scores $s_i = q K_i^\top/\sqrt{d}, i\in [T]$, we define the \textbf{(query-adaptive) oracle criterion}  $\oraclecriterion$ which takes the head's scores $s$ as input:
\begin{equation}
\label{eq:oracle}
    \oraclecriterion(s) = \indicator{ \frac{\sum_{i\in \mathcal I} \exp(s_i)}{\sum_{i \in \mathcal I} \exp(s_i)  +\sum_{i \notin \mathcal I} \exp(s_i) } \geq \thrsoracle}.
\end{equation}
If the criterion is satisfied for a given query, that is, if $\oraclecriterion=1$, the head \textit{mostly attends} to tokens from the \textit{local window}, and we call it a \textit{local} head. On the other hand, if  $\oraclecriterion=0$, the head assigns at least $1-\thrsoracle$ attention mass to tokens from the global context, and we call it \textit{long-context}. Note that our oracle criterion requires the computation of all the head's attention scores--as such, it is a tool of analysis, but it cannot be used as a practical way to increase compute efficiency.

\section{Method}
\label{sec:method}
Given that many attention heads swing between being local and being long-context depending on the input token (as illustrated in Figure~\ref{fig:gaussian} and further observed in Section~\ref{sec:downstream_analysis}), how can we identify local heads in a query-adaptive manner while only computing the attentions scores from the local window? Intuitively, we want a criterion that can distinguish between the two following cases:

\begin{itemize}
    \item \textit{Case 1 (long-context head):} The scores from the local window follow the same distribution as the remaining scores (second plot in Figure~\ref{fig:scores}), and thus tokens from the local window cannot make up for most of the mass. 
    
    \item \textit{Case 2 (local head):} The scores from local tokens are significantly ``out-of-distribution'' on the right-sided tail (first plot in Figure~\ref{fig:scores}). While this does not guarantee that the attention head assigns most of the mass to those tokens, as there might be outliers in the distribution of the non-local scores (third plot in Figure~\ref{fig:scores}),
    this motivates us to label the head as a local head.
\end{itemize}

But how can we efficiently distinguish between the two cases? The key insight is that a Gaussian approximation for the keys, which in turn yields a  Gaussian approximation for the scores (black dashed line in Figure~\ref{fig:scores}), provides a good approximation for deciding what is ``in-distribution'' (Case 1) and what is ``out-of-distribution'' (Case 2). Such an approximation in turn allows us to construct an efficient approximate version of the oracle criterion from Equation~\eqref{eq:oracle}, that we call \textbf{query-adaptive attention (QAdA)}.

\begin{figure*}[t]
\begin{lstlisting}[language=Python, caption=Query-adaptive attention (QAdA) with local window approximation, label={lst:adaptive_attention}],float=th]
def adaptive_attention(q, k, v, mean_k, cov_k, Tl=128, log_thrs=0.6):
    mean_s = einsum('bhnd,hd->bhn', q, mean_k), 
    var_s = einsum('bhnd,hde,bhne->bhn', q, cov_k, q)
    numerator = lse(q @ k[:,:, local_indices]/sqrt(d), dim=-1)
    log_bulk = log(seq_len - window_size) + var_s / 2 + mean_s
    denominator = lse(stack([numerator, log_bulk]),dim=0)
    mask = numerator - denominator > log(log_thrs)
    out[mask], out[!mask] = local_attn_(q, k, v,  mask), dense_attn_(q, k, v,  !mask)
    return out
\end{lstlisting}
\end{figure*}

\subsection{Query-adaptive criterion}
The computational bottleneck in the oracle criterion from Equation~\eqref{eq:oracle} arises from the un-normalized mass $A^{\text{bulk}} := \sum_{i \notin \mathcal I} \exp(s_i)$ of the tokens from the bulk (see Figure~\ref{fig:gaussian}).  Let $\nu^{\text{bulk}}$ be the \textit{empirical distribution} of the keys $k_i^\top$, $i \in [T] \setminus \mathcal I$ and let $\tbulk=T- T_{\text{local}}$. We can write the un-normalized mass as an expectation over $\nu^{\text{bulk}}$:
\begin{equation}
    A^{\text{bulk}}= \tbulk~\mathbb E_{k^\top \sim \nu^{\text{bulk}}} \exp\left(\frac{qk_i^\top }{\sqrt{d}}\right). 
\end{equation}
The main idea behind our method is to now approximate $\nu^{\text{bulk}}$ by a product of Gaussians distributions with some mean $\mu_K$ and covariance $\Sigma_K$ (defined in Section~\ref{sec:pipeline}):

\begin{equation}
    \nu^{\text{bulk}} \approx \left(\mathcal  N(\mu_K, \Sigma_K)  \right)^{\tbulk}.
\end{equation}
Such an approximation clearly does not apply at the level of individual keys. Indeed, according to the Gaussian approximation, all keys should be identically distributed. However, this is definitely not the case as any two distinct keys store different positional information. Nevertheless, when averaged over the keys, we can hope that on a macro distribution level the approximation is accurate. More precisely, we propose to approximate:
\begin{align}
    \underset{k^\top \sim \nu^{\text{bulk}}}{\mathbb E} \exp\left(\frac{qk^\top}{\sqrt{d}}\right) &\approx \underset{k^\top \sim \mathcal N(\mu_K, \Sigma_K)}{ \mathbb{E}} \exp\left(\frac{q k^\top}{\sqrt{d}}\right). \label{eq:gaussianapprox}
\end{align}
In fact, the RHS can be computed in closed form. Indeed, we note that $\exp(q k^\top/\sqrt{d})$ follows a log-normal distribution:
\begin{align}
        \underset{k^\top \sim \mathcal N(\mu_K, \Sigma_K)}{ \mathbb{E}} \exp\left(\frac{q k_i^\top}{\sqrt{d}}\right) &=   \underset{s\sim \mathcal N(\mu_s, \sigma^2_s)}{ \mathbb{E}} \exp(s)  \nonumber\\
        &= \exp(\mu_s + \sigma_s^2/2)
        \label{eq:approx}
\end{align}
with $\mu_s = q \mu_K^\top/\sqrt{d}$ and $\sigma^2_s = q \Sigma_K q^\top /d$ the mean and variance of the scores. 
Assuming that we have access to the mean $\mu_K$ and covariance $\Sigma_K$ statistics (see Section~\ref{sec:moments}), we can therefore compute an approximation of $ A^{\text{bulk}}$ in constant run-time wrt.~$T$! 

 In summary, given the moments $\mu_K$ and $\Sigma_K$, the query $q$ and  the  scores $s_i$ obtained from the local keys $k_i$, $i \in \mathcal I$, we propose to approximate  the oracle criterion in Equation~\eqref{eq:oracle} via the following query-adaptive criterion (QAdA) with $A^{\text{local}} = \sum_{i\in \mathcal I} \exp(s_i)$:
\begin{align}
 c^h_{\text{approx}}(s)=
        \indicator{ \frac{A^{\text{local}}}{ A^{\text{local}} +\tbulk \exp\left(\mu_s +  \sigma_s^2/2 \right) } 
        \geq \thrsapprox}
        \label{eq:adaptive_criterion}
\end{align}

\subsection{Computing $\mu_K$ and $\Sigma_K$}
\label{sec:moments}
 
\underline{\textit{Option 1 (current prompt):}} After pre-filling and before generation, we can compute the moment statistics from the current KV-cache. That is, we compute $\mu_K = \frac{1}{T^{\text{bulk}}}\sum_{i \in [T] \setminus \mathcal I} K_i $ and $\Sigma_K =\frac{1}{T^{\text{bulk}}}\sum_{ i \in [T] \setminus \mathcal I} K_i K_i^\top - \mu_K \mu_K^\top $. As a result, the moment statistics capture information from the keys contained in the bulk. 
A key point to note is that while the definition of $\mu_K$ involves a sum over all the bulk tokens, computing $\mu_{K}$ does \textbf{not} cost $O(Td)$ operations per token, as it can be updated at each step during decoding for a cost of $O(d)$ operations by using the fact that $\mu_{K, T+1} =  \frac{1}{T^{\text{bulk}} + 1}\sum_{i \in [T+1] \setminus \mathcal I} K_i = 
\frac{T^{\text{bulk}}}{T^{\text{bulk}} + 1}\mu_{K, T} + \frac{K_{T+1}}{T^{\text{bulk}} + 1}$.
The same applies to $\Sigma_k$ (for an update cost of  $O(d^2)$ operations).

\underline{\textit{Option 2 (other prompt):}} Maybe surprisingly, we show in Section~\ref{sec:ablations}  and Appendix~\ref{sec:additional_exps} that we obtain more robust performances by computing the mean $\mu_K$ and covariance $\Sigma_K$ from keys generated from a \textit{different prompt} of similar length. We refer the reader to Appendix~\ref{sec:keys} for additional details. While such an approach may appear counter-intuitive, we hypothesize that $\mu_K$ and $\Sigma_K$ benefit from reflecting a ``generic distribution of keys'', rather than that of the current prompt. While the underlying reasons for this remain unclear, this intuition is supported by the fact that we show in Section~\ref{sec:ablations} that using a \textit{random words} prompt yields robust performance.  While the distribution of keys becomes independent of the current prompt, query-dependency still persists as inner product involves the query.

\begin{figure*}[t]
\centering
\includegraphics[width=\linewidth]{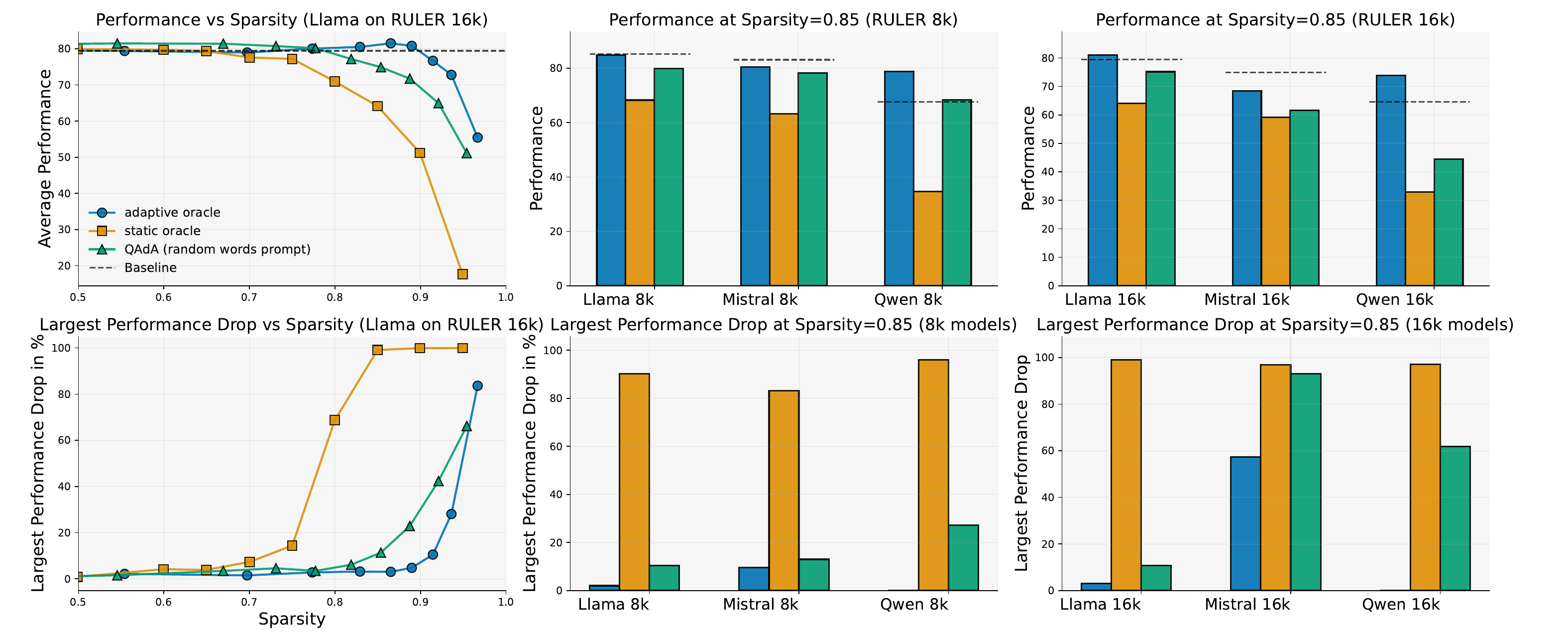}
    \caption{\small Comparison of QAdA against the adaptive and static oracles on the RULER benchmark.  
\textit{Left:} For Llama 3-8B, we show the average performance (top) over the selected RULER 16k tasks as a function of the average sparsity for varying thresholds $\tau$, along with the worst-case performance drop (\%) compared to the baseline performance among the selected tasks.  
\textit{Middle and Right:} Average performance and worst-case drop for a fixed sparsity level of 0.85 across three model families—Llama, Mistral, and Qwen—on RULER 8k (center) and RULER 16k (right).  
Our adaptive criterion consistently matches or outperforms the static oracle criterion, and in some cases (e.g., Mistral), even achieves performance comparable to the adaptive oracle.}
    \label{fig:compare-approx}
\end{figure*}

\subsection{Summary of inference pipeline and run-time complexity}
\label{sec:pipeline}

We describe how our adaptive criterion can be applied in practice by decoding LLMs and explain how this can lead to decreased run-time complexity.

Before starting generation,  we calculate the moment statistics $\mu_K$ and $\Sigma_K$. Then, during decoding, before computing the attention output for a layer, we update the moment statistics $\mu_K$ and $\Sigma_K$ and apply the query-adaptive criterion to every head in the layer, thus labeling a subset of them as local heads. We approximate the output of those using a local window, and compute the output of the others the usual way. We summarize the procedure in Listing~\ref{lst:adaptive_attention}.

\begin{table}[t]
    \centering
    \begin{tabular}{lll}
        \toprule
        criterion & criterion comp. & attention comp. \\
        \midrule
        none&  -  & $O(Td)$ \\
        oracle &  $O(Td)$ & $O((1-\rho) Td +  \rho T_{\text{local}}d)$ \\
        QAdA &  $O(T_{\text{local}}d +d^2)$  &  $O( (1-\rho) Td + \rho T_{\text{local}}d)$  \\
        \bottomrule
    \end{tabular}
        \caption{\small Run-time complexity of the oracle and adaptive criterion, as well as the cost of computing the resulting approximate attention. $\rho$ is the fraction approximated by a local window of size $T_{\text{local}}$. }\vspace{-0.2in}
    \label{tab:complexity}
\end{table}

Unlike the oracle criterion from Equation~\eqref{eq:oracle}, our query-adaptive criterion achieves a constant run-time complexity in $T$ assuming that $T_{\text{local}} \ll T$. 
Moreover, let $\rho$ be the fraction of times a head has been labeled as local head and $d$ be the head dimension: then the average cost of computing the next token using the (approximated) attention mechanism  is $O((1-\rho)  Td + \rho  T_{\text{local}}d)$, as opposed to the $O(Td)$ operations required by the standard attention mechanism. These computations are summarized in Table~\ref{tab:complexity}.

\begin{figure*}[t]
\centering
\includegraphics[width=\linewidth]{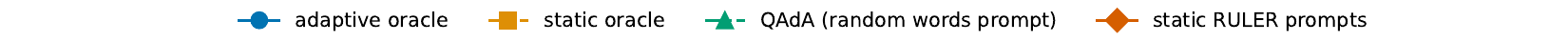}
    \begin{subfigure}[b]{0.24\linewidth}
        \centering
\includegraphics[width=\linewidth]{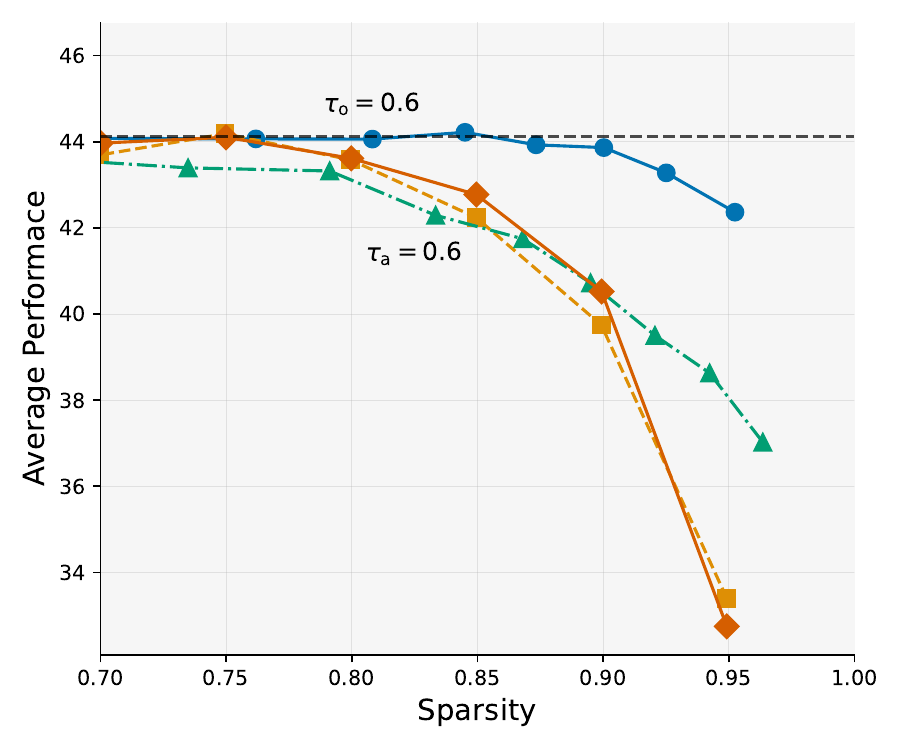}
        \caption{LongBench Performance}
        \label{fig:longbench}
    \end{subfigure}
    \begin{subfigure}[b]{0.24\linewidth}
        \centering
\includegraphics[width=\linewidth]{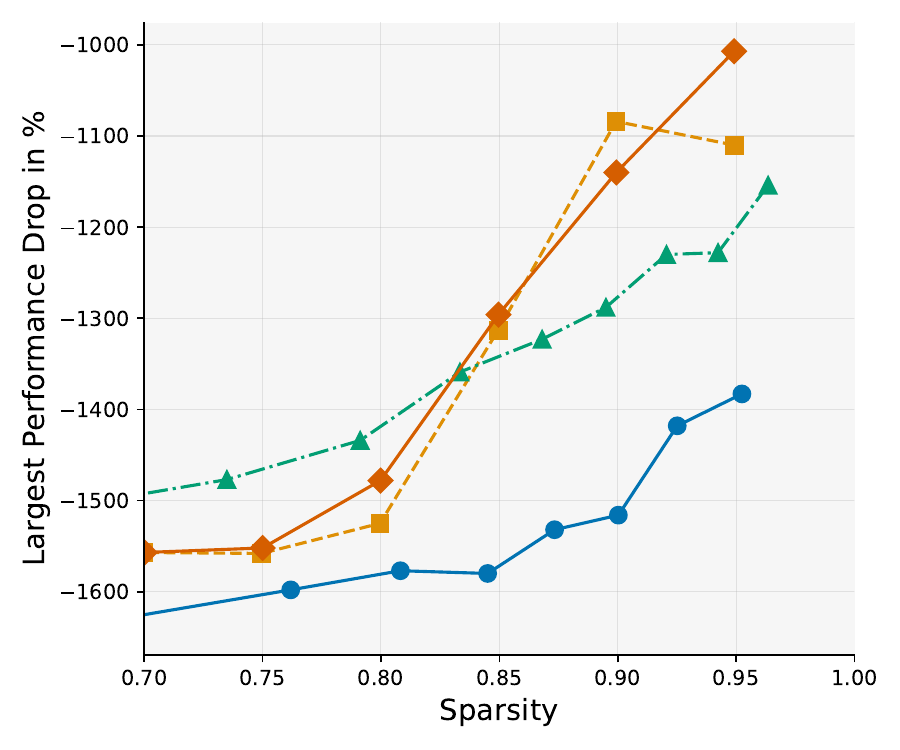}
        \caption{LongBench Max Drop}
        \label{fig:longbenchmin}
    \end{subfigure}
    \begin{subfigure}[b]{0.24\linewidth}
        \centering\includegraphics[width=\linewidth]{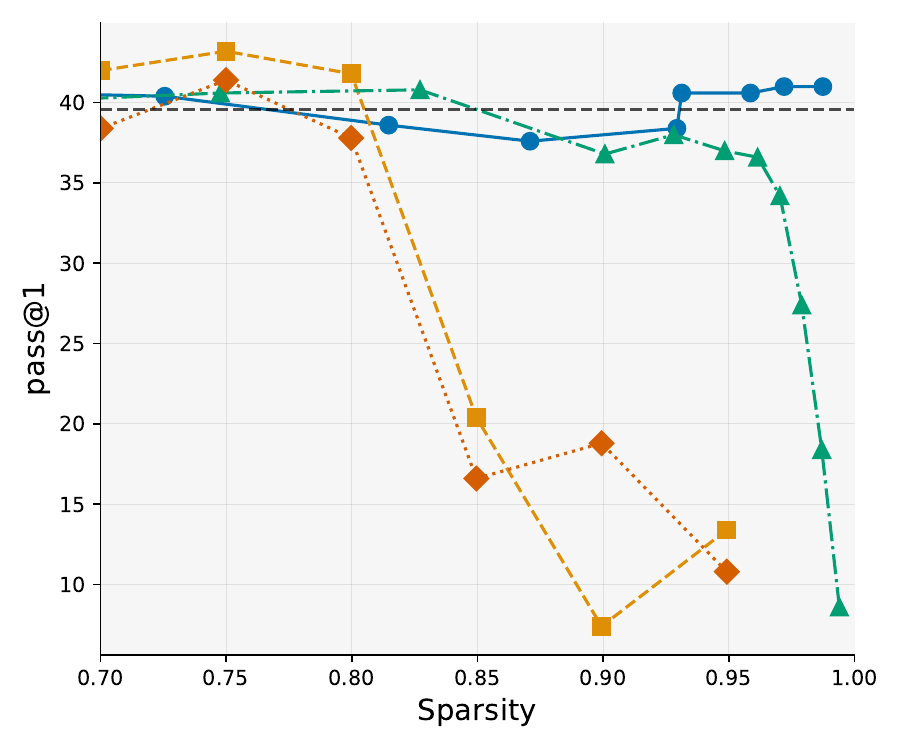}
        \caption{Long-Context MBPP}
        \label{fig:mbpp}
    \end{subfigure}
    \begin{subfigure}[b]{0.24\linewidth}
        \centering\includegraphics[width=\linewidth]{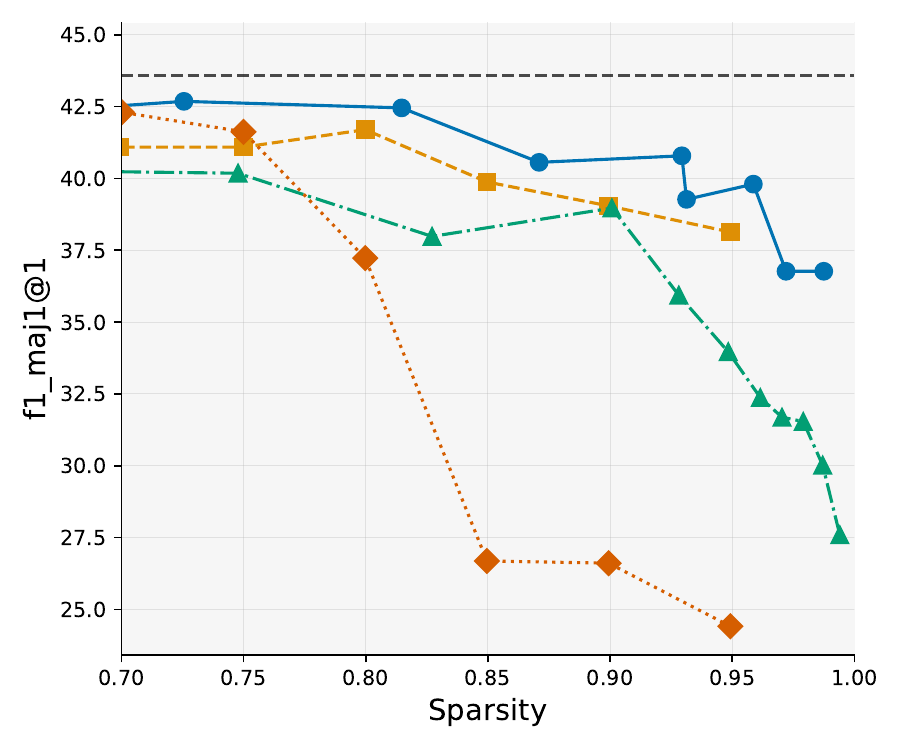}
        \caption{Long-Context GSM8k}
        \label{fig:gsm8k}
    \end{subfigure}
        \begin{subfigure}[b]{0.64\linewidth}\centering
        \includegraphics[width=0.8\linewidth]{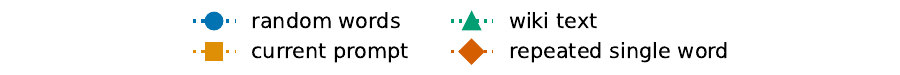}
        \begin{subfigure}[b]{0.49\linewidth}
        \includegraphics[width=\linewidth]{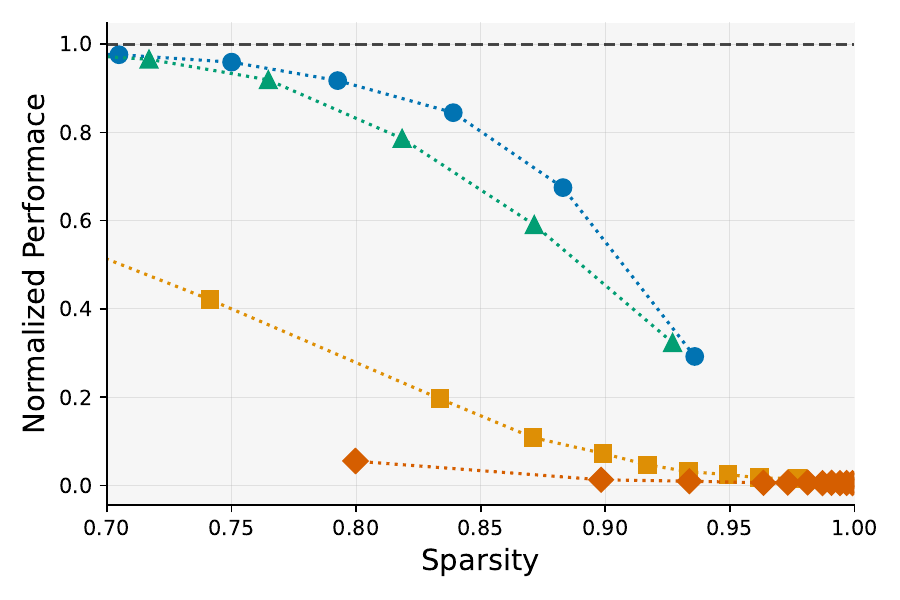}
                            
        \caption{varying prompts, ``vt'' task}
        \label{fig:prompts}
        \end{subfigure}
        \begin{subfigure}[b]{0.49\linewidth}
        \includegraphics[width=\linewidth]{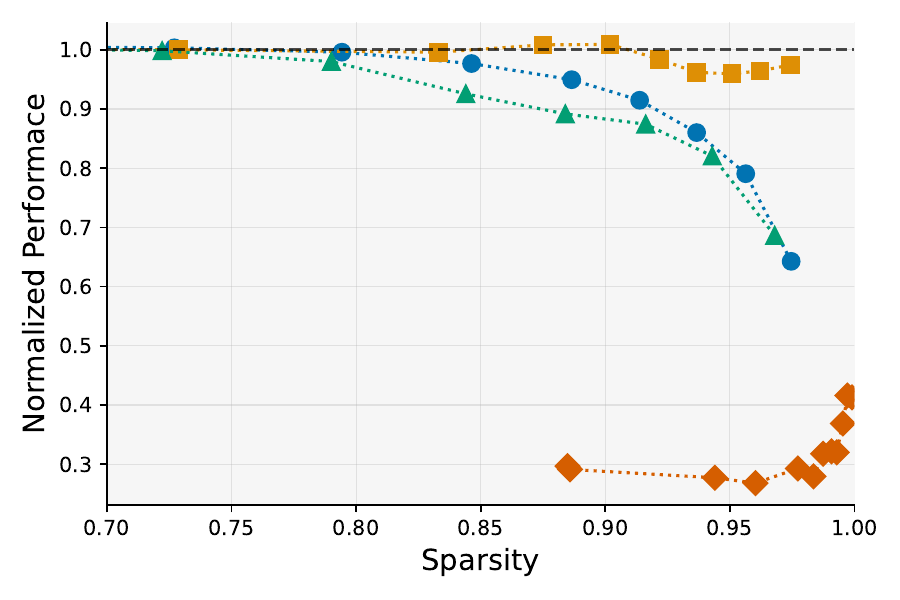}
        \caption{varying prompts, ``fwe'' task}
        \label{fig:promptsfwe}
        \end{subfigure}
    \end{subfigure}
\begin{subfigure}[b]{0.32\linewidth}
        \centering
        \includegraphics[width=\linewidth]{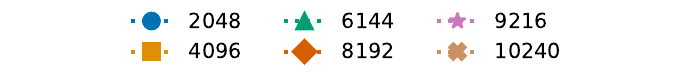}
        \includegraphics[width=\linewidth]{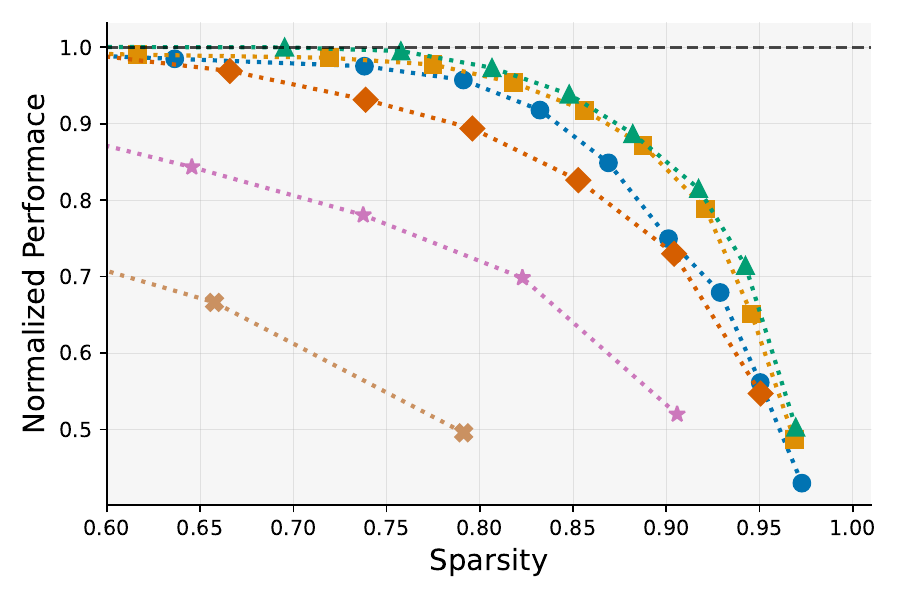}
        \caption{varying seq. lengths}
        \label{fig:seqlen_prompt}
        \end{subfigure}

    \caption{\small \textbf{Top row:} Similar to Figure~\ref{fig:compare-approx}, we show the average  performance for the LongBench benchmark, the pass@1 score for the MBPP task and the f1-score for the GSM8k task. 
    \textbf{Bottom row:}   Ablations for the content of the prompt (e-f) and the length of the prompt (g) used to generate the mean $\mu_K$ and covariance $\Sigma_K$ for the adaptive criterion from Section~\ref{sec:method}.  We show the normalized performance as a function of sparsity  (e) for the  ``vt'' task and (f) for the ``fwe'' task and (g) averaged over the RULER $8$k tasks,  respectively.}    
\end{figure*}

\section{Evaluation on downstream tasks}
\label{sec:downstream}

\subsection{Experimental Setting}
\label{sec:expsetting}

\paragraph{Datasets.}
We evaluate on the two standard long-context benchmarks, RULER \citep{hsieh2024ruler} and LongBench \citep{bai2024longbench}. We also propose long-context variants of  GSM8k \citep{cobbe2021training} and MBPP \citep{austin2021program}, where we  ``hide''  informative few-shot examples in a long-context prompt containing roughly $\approx 10k$ tokens. We refer the reader to Appendix~\ref{sec:apx-exp-details} for further details.

\paragraph{Models.}
Our default model is the instruction fine-tuned \textit{Llama 3-8B} model.
We also use the two models
 \textit{Mistral-7B-Instruct-v0.2} and  \textit{Qwen2-7B-Instruct} as provided by \textit{HuggingFace}.
    To account for longer contexts, we set our models' RoPE parameter to $\theta = 2'000'000$, which is approximately the value from the NTK-aware interpolation \citep{peng2023ntk} for a context length of $32k$. 
 For all evaluations, we choose a temperature of $0$, i.e.~use the greedy decoding strategy. We always let $T_{\text{local}} = 128$ and use the first $T_{\text{init}}=16$ tokens as ``attention sink'' tokens, leaving $112$ tokens from the neighorhood closest to the current token (or sliding window).

\paragraph{Methods}
We implement the query-adaptive \textbf{oracle} criterion (Equation~\ref{eq:oracle}), alongside with two query-independent static criteria, \textbf{static oracle} and \textbf{static RULER}. The static method, for a fixed sparsity threshold of $\alpha$ (we ablate over intervals of 5\%), permanently labels as local the $\alpha$ percentage of heads that were most often labeled as local by the oracle criterion on prompts from the RULER tasks. 
The oracle static method, for a fixed sparsity threshold of $\alpha$, labels as local the $\alpha$ of heads that are most often labeled as local by the oracle criterion on the prompts of the processed task. See Appendix~\ref{sec:apx-exp-details} for further details.

We implement QAdA from Section~\ref{sec:method} for four choices of prompts (see Section~\ref{sec:moments}): The current prompt, described as Option 1 in Section~\ref{sec:moments}, and three variants of Option 2:  randomly sampled independent words from Wikipedia (\textit{random words prompt}), concatenated Wikipedia extracts (\textit{wiki prompt}), and repetitions of single word (\textit{single word prompt}). Only the statistics $\mu_K$ and $\Sigma_K$ generated from  the current prompt contain information about the prompt, while the others are agnostic to the current prompt. Our ablation in Subsection~\ref{sec:ablations} suggest that Option 2 (random words prompt) yields the most robust performance. 

\paragraph{Metrics}
We use the standard metrics for evaluation provided by the corresponding benchmarks, which we refer to as the \textbf{performance}. For the LongBench benchmark, we compute the average normalized performance \textbf{(avg. norm. performance)}, which is obtained by dividing the performance by the performance of the standard full attention model. We always plot the performance as a function of the \textbf{sparsity}, that is the average percentage of heads labeled as local heads, and thus approximated by a local window. For both our adaptive, as well as the static criteria, the sparsity almost directly translates into a reduction of FLOPs used by the attention mechanism (minus a small constant overhead to compute the local scores).

\begin{figure*}
    \centering    
        \begin{subfigure}[b]{0.29\linewidth}
        \centering
\includegraphics[width=\linewidth]{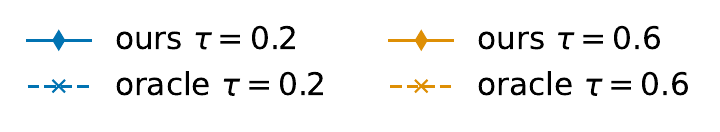}\\
\includegraphics[width=\linewidth]{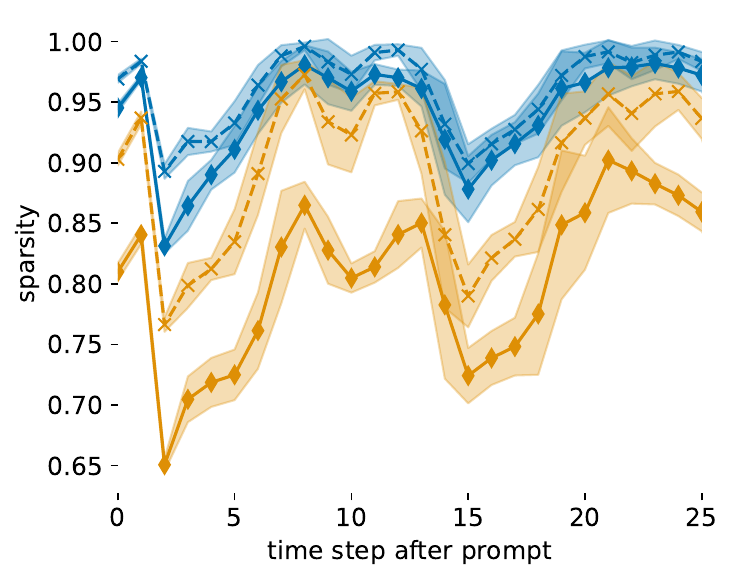}
        \caption{query-wise adaptivity}
        \label{fig:timewise-fwe}
    \end{subfigure}
\begin{subfigure}[b]{0.34\linewidth}
        \centering
        \includegraphics[width=\linewidth]{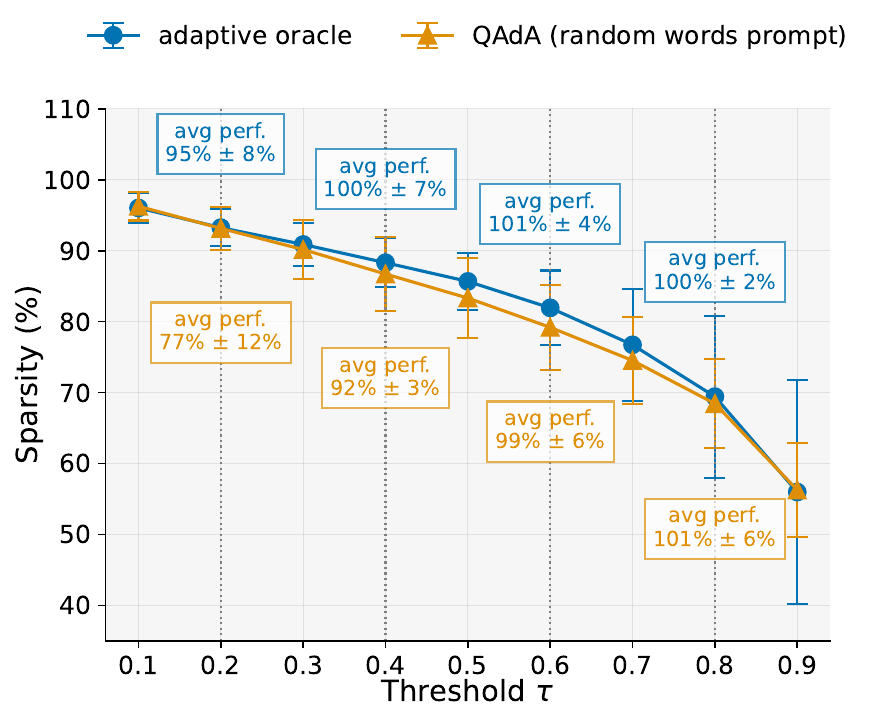}
        \caption{oracle vs QAdA}
        \label{fig:threshold}
    \end{subfigure}
\begin{subfigure}[b]{0.34\linewidth}
        \centering
        \includegraphics[width=\linewidth]{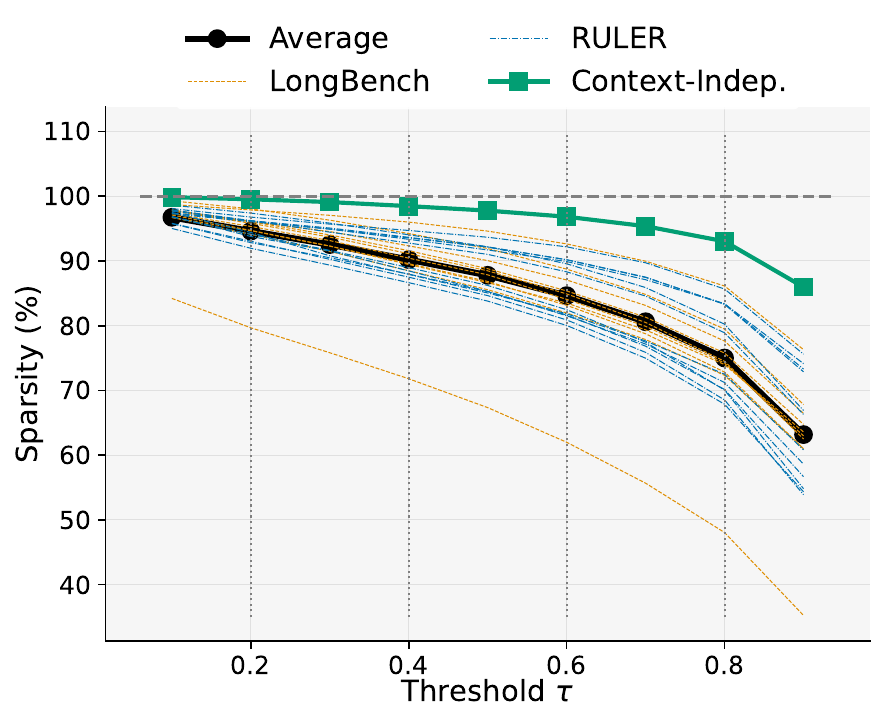}
        \caption{sparsity of QAdA by tasks}
        \label{fig:threshold_per_task}
    \end{subfigure}
    \caption{\small 
    \textbf{a)} The mean and standard deviation of the fraction of heads labeled as local heads as a function of time-steps for prompts from the ``fwe'' task. 
   \textbf{ b)}  The average sparsity and standard deviation as a function of the threshold $\tau$ for Llama 3-8B over the RULER 8k and 16k, as well as the LongBench tasks. The annotations show the mean and standard deviation of the normalized performances (with $1$ being the performance of the standard dense attention). 
\textbf{c)} The average sparsities as a function of the threshold $\tau$, similar to those shown in b), are presented for each task, specifically for the QAdA criterion. Additionally, we present the average sparsity for a context-independent task. This task does not require context to be solved, and we observe that QAdA labels significantly more heads as local heads for the same threshold.} 
\end{figure*}

\vspace{-0.1in}
\subsection{Performance on RULER and LongBench}
\label{sec:downstream_analysis}
\paragraph{Oracle gains over static.} We begin by comparing the adaptive oracle criterion against the  static oracle criterion. We observe significant gains in performance across all models on the RULER benchmark in Figure~\ref{fig:compare-approx}, both in terms of the average performance, as well as the worst-case performance drop. The same observation also holds for the experiments on the LongBench benchmark in  Figure~\ref{fig:longbench},\ref{fig:longbenchmin}. For instance, for the Llama model we see a 20\% increase in sparsity on the RULER tasks (from $\approx 70\%$ to $\approx 90\%$) and a $\approx 5-10\%$ increase on LongBench tasks at fixed performance level. These results underline the potential gains that are achievable by adaptive criteria for selecting attention heads over static ones.

\paragraph{QAdA outperforms static.}
We observe that our efficient adaptive criterion significantly outperforms the static criterion  on the RULER task for sequence lengths of $8$k in Figure~\ref{fig:compare-approx}, and also for lengths $16$k for the Llama model. 
Moreover, our adaptive criterion matches the performance of the oracle static criterion and even slightly outperforms it on LongBench in Figure~\ref{fig:longbench} and Mistral on RULER $16$k. The only situation where we see performance drops compared to the static method is for Qwen on RULER $16$k, where the score of the baseline model is itself very low. These results demonstrate that our criterion is capable of exploiting the query-adaptivity of attention heads.

\paragraph{Outperforming the standard dense attention with Qwen} Finally, we observe in Figure~\ref{fig:compare-approx} that both the oracle adaptive criterion and our adaptive criterion surpass the baseline performance of the standard full attention for Qwen on RULER $8k$ (see Figure~\ref{fig:qwen9k} in the Appendix).
These gains are even more visible for the oracle criterion on RULER $16$k, where we find an average performance increase of more than 15 points for a sparsity of $0.85$.  
It is also worth noting that these gains are made possible by a query-adaptive approach and do not occur for static methods. These improvements highlight the fact that in long-context settings, models may attend to unnecessary parts of the context, which the query-adaptive criterion can effectively prune. Consequently, in such settings, the query-adaptive criterion can provide benefits beyond computational efficiency, also leading to enhanced performance.

\subsection{Performance on reasoning and code tasks }
While both the RULER and LongBench benchmarks require only short answers (sometimes less than 20 tokens), we also wonder how well our method is capable of selecting the ``right'' heads in challenging reasoning and code generation tasks, where the expected answers tend to be longer. 
We propose two long-context variants of the GSM8k and MBPP tasks (we provide examples in the Appendix) where we hide a few relevant few-shot examples in a mostly irrelevant long prompt. As instruction fine-tuned models do not require few-shot COT examples for solving the tasks, we instead use the pre-trained version of Llama 3-8B which heavily relies on these examples.

We show in Figure~\ref{fig:mbpp} and Figure~\ref{fig:gsm8k} the performances on the long-context variants of the two tasks as a function of sparsity. We again observe that our adaptive criterion yields robust performance, outperforming the static criteria. Particularly striking are the gains for the long-context MBPP task, where both the oracle criterion and our query-adaptive criterion let us approximate almost all heads as local heads (more than 95\%), while the performance of the static approaches significantly decreases beyond $80\%$ sparsity.

\subsection{Ablations over moment statistics}
\label{sec:ablations}
In this section, we present  ablations for the choice of the prompt used to generate the  mean $\mu_K$ and variance $\Sigma_K$ statistics, as described in Section~\ref{sec:moments}.

\paragraph{Prompt.} We ablate in Figure~\ref{fig:prompts}-\ref{fig:promptsfwe} over the content of the prompts used to generate the moments statistics. We show the curves only for the two illustrative RULER tasks ``variable tracing'' (``vt''), that has a highly repetitive structure,  and ``frequent word extraction'' (``fwe'').
Maybe surprisingly, we find for the ``vt'' task that the best performance is attained when using randomly sampled words, while repetitively using the same words results in the worst performance. Moreover, using the exact moments (i.e., \textit{current prompt}) also results in very poor performance. This is not the case for the ``fwe'' task, where using the current prompt achieves the best performance. We believe that the failure on the ``vt'' task is explained by the repetitive structure of the prompt, which resembles the structure of the repeated single word prompt that also yields very poor performance. 
 In summary, we find that although using  ``current prompt'' can sometimes yield strong performance (``fwe'') task, it is not robust to the choice of task. In contrast,  ``random words prompt'' using a distinct dataset yields more robust performance. We present additional related experiments in Figure~\ref{fig:ablations-vt-extra} in the Appendix.

\paragraph{Sequence Length.}
We compare in Figure~\ref{fig:seqlen_prompt} the performances of our query-adaptive method using Option 2 (random words prompt) for different lengths of the prompt used to generate the  mean $\mu_K$ and covariance $\Sigma_K$.  We show the average normalized performance across all RULER  $8$k tasks.
We see drastic drops in performance when the prompt used to compute the statistics gets longer than the length of the actual prompt (that is $\approx 8100$ tokens long), whereas performance is surprisingly robust to variations for shorter sequence lengths. This dependence to the length of the random words prompt suggests that while the statistics $\mu_K$ and $\Sigma_K$ do not contain any information about the task (as we use random words), they nevertheless contain positional information critical for the criterion to identify the right set of local heads.

\section{Discussion: Adaptivity to contexts}
\label{sec:rec}
We saw in the previous section that QAdA is capable of selecting relevant heads for solving the corresponding long-context tasks. In this section, we investigate which heads are selected by the model, and to what extent the model selects heads based on the context. Besides prompts from the RULER and LongBench tasks, we also study the behavior on a \textit{context-independent} task where. More precisely, we take the context from the ``qa-2'' task from the RULER benchmark but replace the question with: \textit{Can you describe LLMs in a few sentences?}. To solve this task, the model does not need to attend on the context, and we show that the model indeed labels more heads as local heads. This shows that the model is capable of \textit{adapting to the context}.

\paragraph{Query-wise sparsity.} As a first question, we investigate whether QAdA is capable of changing the sparsity (average fraction of heads labeled as local heads) on a query-wise basis. We provide an illustrative example in Figure~\ref{fig:timewise-fwe}, showing the average percentage of heads chosen by both the oracle and the adaptive criterion as a function of the time-step (query). We choose the "fwe" task, for which all responses to the prompts follow exactly the same pattern, and plot the mean and standard deviation as a function of the index of the generated token.
We observe that the trend of our adaptive criterion aligns closely with the trend of the oracle criterion, and both vary strongly from token to token.

\paragraph{Sparsity vs. Threshold.} We further plot in Figure~\ref{fig:threshold} the average sparsity and the standard deviation of QAdA and the oracle criterion as a function of the threshold $\tau$.
We make two findings: first, that QAdA closely follows the sparsity of the adaptive oracle criterion but tends to label slightly more heads as long-context. Second, that the standard deviation of the average sparsity (with respect to different tasks) is non-negligible, meaning that the sparsity can vary from task to task. This indicates that our adaptive criterion effectively adjusts the level of sparsity and is capable of adapting to "difficult" tokens. Indeed, we further show in Figure~\ref{fig:threshold_per_task} the average sparsities for each task for QAdA. We also plot in green the average sparsity when asking the model to generate a response for a task that does not require any knowledge from the context. As we can see, the QAdA uses significantly fewer heads as long-context heads for this task than for the other tasks at the same threshold.

\begin{figure*}
    \centering    
        \begin{subfigure}[b]{0.49\linewidth}
        \centering
\includegraphics[width=\linewidth]{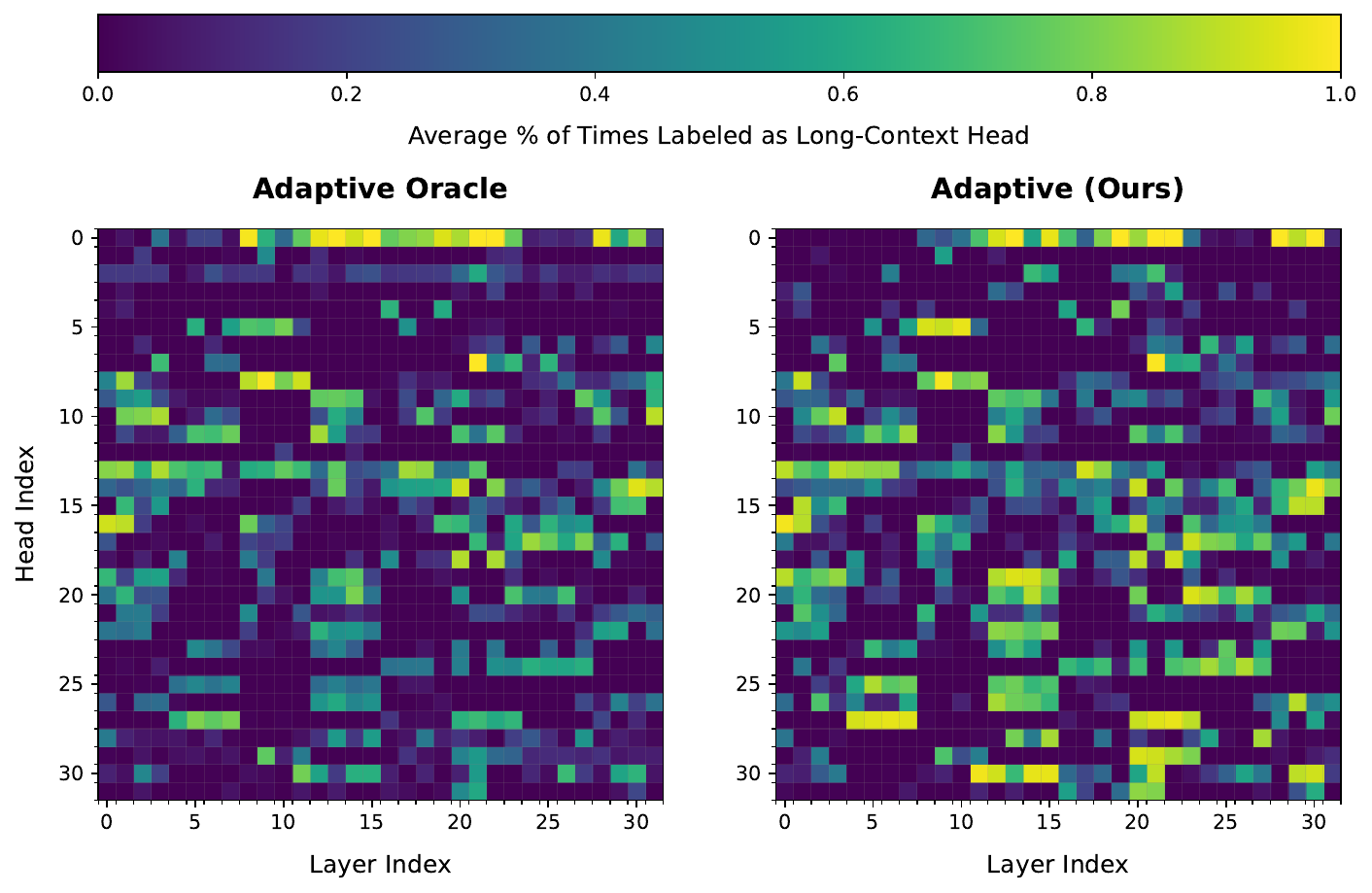}
        \caption{RULER tasks}
        \label{fig:heatmap}
    \end{subfigure}
        \begin{subfigure}[b]{0.49\linewidth}
        \centering
\includegraphics[width=\linewidth]{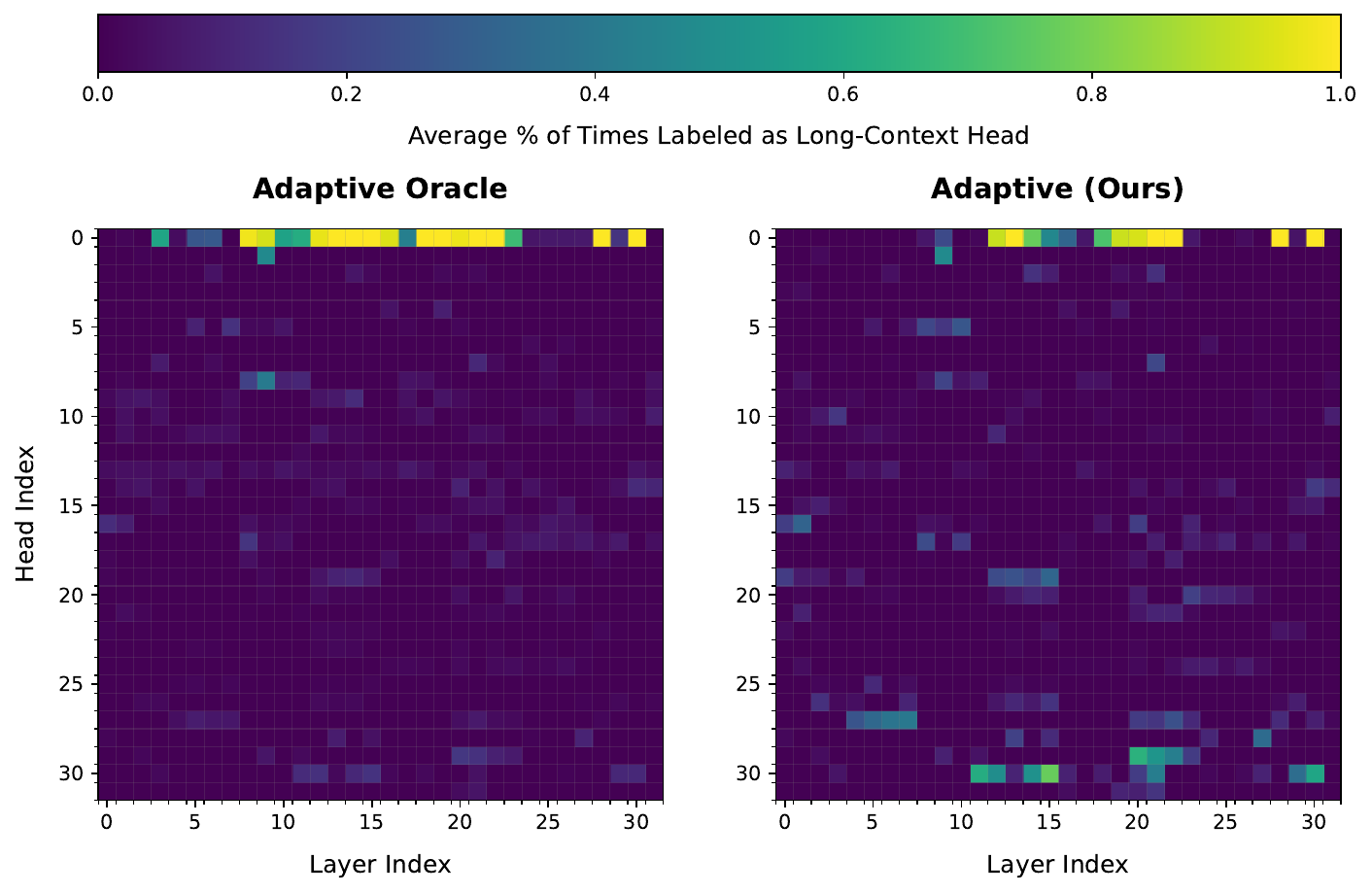}
        \caption{context-independent task}
        \label{fig:heatmap_independent}
    \end{subfigure}
    \caption{\small  We show for both the oracle adaptive and the adaptive criterion the \% of times each head has been labeled as long-context head averaged over a) the six RULER 8k tasks with  $\thrsoracle=\thrsapprox=0.6$ and b) the context-independent task based on the ``qa-2'' task from RULER.} 
    \label{fig:heatmaps}
\end{figure*}

\paragraph{Distribution of local heads across layers.}
Finally, in Figure~\ref{fig:heatmap} and Figure~\ref{fig:heatmap_independent}, we show the average percentage of times each head has been labeled as long-context for the RULER tasks and the context-independent tasks. For the RULER tasks, which require the model to look at the entire context, we see that both criteria show matching patterns and long-context heads occur across all layers. This demonstrates that our adaptive criterion successfully identifies long-context heads across all layers. Moreover, for the context-independent task, we see that while the first layer still attends to the full context, all layers are essentially always approximated by the local windows.

\section{Related works}
\label{sec:relatedworks}
There is an overwhelming body of work studying and exploiting sparsity in attention heads. We refer the reader to \citep{wan2023efficient,zheng2024attention} for surveys and only discuss the most directly related works here.

\paragraph{Static classification of heads } 
\citet{wu2024retrieval} showed that a few attention heads, called ``retrieval heads,'' are particularly critical in retrieving long-context information, with multiple follow-up works \citep{tang2024razorattention,hong2024token,xiao2024duoattention,cai2024pyramidkv,he2025task}. Most related to this paper is \citet{xiao2024duoattention}, who also proposed dividing the heads into long-context and local heads. All these methods statically assign labels to the heads before generation. They do so by analyzing attention patterns on selected tasks, or, as done in \citep{xiao2024duoattention}, learn the assignment using gradient descent. Our paper crucially differs from these works as we explore the \textit{query-adaptive} nature of attention heads to their changing contexts and do not require an additional dataset to label the heads.

\paragraph{Query-adaptive sparsity.}
Similar to this paper, there is an entire line of research that exploits query-dependent sparsity in some way. For instance, numerous works propose efficient approximations that retrieve per head the subset of tokens with the highest scores \citep{tang2024quest,ribar2023sparq,chen2021scatterbrain,sun2024shadowkv}. For context, multiple works also propose static variants that select the tokens for all queries \citep{zhang2023h2o,li2024snapkv,oren2024transformers}. These works are complementary to this paper. More related to this paper is the approach taken by \citep{liu2023deja,akhauri2024shadowllm}, where a classifier is trained to dynamically predict which attention heads can be ``dropped.'' The classifier takes as input the residual of an earlier layer and thus also adapts to the changing contexts. However, our paper crucially differs in two ways: first, we do not rely on any additional dataset for labeling the heads, nor do we require training an additional classifier. Second, we also distinguish between local and long-context heads, and do not simply drop heads.

\section{Limitations}
This paper highlights the query-adaptive nature of attention heads in the way they retrieve long-context information, and provides a second order statistics-based test for locality. However, we do not test and provide a highly optimized implementation compatible with flash attention, and we do not showcase real run-time gains. This was out of scope for the current work and is an exciting area for future research.

\section{Conclusions }
Our first key finding shows that the attention head exhibits two distinct behaviors: local- it attends to local tokens and long-context- it attends to tokens beyond local tokens. This behavior is query-dependent, and perhaps surprisingly, a simple test QAdA (Query-Adaptive Attention) based on the second-order statistics of the keys and local scores is quite effective in predicting this behavior. We tested the efficacy of QAdA through state-of-the-art models such as Llama, Qwen, and Mistral (7 to 8 billion parameters) and various important long-context benchmarks, including RULER and Longbench. Through rigorous ablations, we present a deeper understanding of the inner workings of the test and the attention mechanism.

\bibliography{paper}

\begin{thebibliography}{37}
\providecommand{\natexlab}[1]{#1}
\providecommand{\url}[1]{\texttt{#1}}
\expandafter\ifx\csname urlstyle\endcsname\relax
  \providecommand{\doi}[1]{doi: #1}\else
  \providecommand{\doi}{doi: \begingroup \urlstyle{rm}\Url}\fi

\bibitem[Ainslie et~al.(2023)Ainslie, Lee-Thorp, de~Jong, Zemlyanskiy,
  Lebr{\'o}n, and Sanghai]{ainslie2023gqa}
Joshua Ainslie, James Lee-Thorp, Michiel de~Jong, Yury Zemlyanskiy, Federico
  Lebr{\'o}n, and Sumit Sanghai.
\newblock Gqa: Training generalized multi-query transformer models from
  multi-head checkpoints.
\newblock \emph{arXiv preprint arXiv:2305.13245}, 2023.

\bibitem[Akhauri et~al.(2024)Akhauri, AbouElhamayed, Dotzel, Zhang, Rush, Huda,
  and Abdelfattah]{akhauri2024shadowllm}
Yash Akhauri, Ahmed~F AbouElhamayed, Jordan Dotzel, Zhiru Zhang, Alexander~M
  Rush, Safeen Huda, and Mohamed~S Abdelfattah.
\newblock Shadowllm: Predictor-based contextual sparsity for large language
  models.
\newblock \emph{arXiv preprint arXiv:2406.16635}, 2024.

\bibitem[Austin et~al.(2021)Austin, Odena, Nye, Bosma, Michalewski, Dohan,
  Jiang, Cai, Terry, Le, et~al.]{austin2021program}
Jacob Austin, Augustus Odena, Maxwell Nye, Maarten Bosma, Henryk Michalewski,
  David Dohan, Ellen Jiang, Carrie Cai, Michael Terry, Quoc Le, et~al.
\newblock Program synthesis with large language models.
\newblock \emph{arXiv preprint arXiv:2108.07732}, 2021.

\bibitem[Bai et~al.(2023)Bai, Bai, Chu, Cui, Dang, Deng, Fan, Ge, Han, Huang,
  et~al.]{bai2023qwen}
Jinze Bai, Shuai Bai, Yunfei Chu, Zeyu Cui, Kai Dang, Xiaodong Deng, Yang Fan,
  Wenbin Ge, Yu~Han, Fei Huang, et~al.
\newblock Qwen technical report.
\newblock \emph{arXiv preprint arXiv:2309.16609}, 2023.

\bibitem[Bai et~al.(2024)Bai, Lv, Zhang, Lyu, Tang, Huang, Du, Liu, Zeng, Hou,
  Dong, Tang, and Li]{bai2024longbench}
Yushi Bai, Xin Lv, Jiajie Zhang, Hongchang Lyu, Jiankai Tang, Zhidian Huang,
  Zhengxiao Du, Xiao Liu, Aohan Zeng, Lei Hou, Yuxiao Dong, Jie Tang, and
  Juanzi Li.
\newblock {L}ong{B}ench: A bilingual, multitask benchmark for long context
  understanding.
\newblock In \emph{Proceedings of the 62nd Annual Meeting of the Association
  for Computational Linguistics (Volume 1: Long Papers)}, pages 3119--3137,
  Bangkok, Thailand, August 2024. Association for Computational Linguistics.
\newblock \doi{10.18653/v1/2024.acl-long.172}.
\newblock \url{https://aclanthology.org/2024.acl-long.172}.

\bibitem[Cai et~al.(2024)Cai, Zhang, Gao, Liu, Liu, Lu, Xiong, Dong, Chang, Hu,
  et~al.]{cai2024pyramidkv}
Zefan Cai, Yichi Zhang, Bofei Gao, Yuliang Liu, Tianyu Liu, Keming Lu, Wayne
  Xiong, Yue Dong, Baobao Chang, Junjie Hu, et~al.
\newblock Pyramidkv: Dynamic kv cache compression based on pyramidal
  information funneling.
\newblock \emph{arXiv preprint arXiv:2406.02069}, 2024.

\bibitem[Chen et~al.(2021)Chen, Dao, Winsor, Song, Rudra, and
  R{\'e}]{chen2021scatterbrain}
Beidi Chen, Tri Dao, Eric Winsor, Zhao Song, Atri Rudra, and Christopher
  R{\'e}.
\newblock Scatterbrain: Unifying sparse and low-rank attention.
\newblock \emph{Advances in Neural Information Processing Systems},
  34:\penalty0 17413--17426, 2021.

\bibitem[Chen et~al.(2024)Chen, Sadhukhan, Ye, Zhou, Zhang, Nolte, Tian, Douze,
  Bottou, Jia, et~al.]{chen2024magicpig}
Zhuoming Chen, Ranajoy Sadhukhan, Zihao Ye, Yang Zhou, Jianyu Zhang, Niklas
  Nolte, Yuandong Tian, Matthijs Douze, Leon Bottou, Zhihao Jia, et~al.
\newblock Magicpig: Lsh sampling for efficient llm generation.
\newblock \emph{arXiv preprint arXiv:2410.16179}, 2024.

\bibitem[Cobbe et~al.(2021)Cobbe, Kosaraju, Bavarian, Chen, Jun, Kaiser,
  Plappert, Tworek, Hilton, Nakano, et~al.]{cobbe2021training}
Karl Cobbe, Vineet Kosaraju, Mohammad Bavarian, Mark Chen, Heewoo Jun, Lukasz
  Kaiser, Matthias Plappert, Jerry Tworek, Jacob Hilton, Reiichiro Nakano,
  et~al.
\newblock Training verifiers to solve math word problems.
\newblock \emph{arXiv preprint arXiv:2110.14168}, 2021.

\bibitem[Dubey et~al.(2024)Dubey, Jauhri, Pandey, Kadian, Al-Dahle, Letman,
  Mathur, Schelten, Yang, Fan, et~al.]{dubey2024llama}
Abhimanyu Dubey, Abhinav Jauhri, Abhinav Pandey, Abhishek Kadian, Ahmad
  Al-Dahle, Aiesha Letman, Akhil Mathur, Alan Schelten, Amy Yang, Angela Fan,
  et~al.
\newblock The llama 3 herd of models.
\newblock \emph{arXiv preprint arXiv:2407.21783}, 2024.

\bibitem[Gu et~al.(2024)Gu, Pang, Du, Liu, Zhang, Du, Wang, and
  Lin]{gu2024attention}
Xiangming Gu, Tianyu Pang, Chao Du, Qian Liu, Fengzhuo Zhang, Cunxiao Du,
  Ye~Wang, and Min Lin.
\newblock When attention sink emerges in language models: An empirical view.
\newblock \emph{arXiv preprint arXiv:2410.10781}, 2024.

\bibitem[He et~al.(2025)He, Liu, and Chen]{he2025task}
Xingyang He, Jie Liu, and Shaowei Chen.
\newblock Task-kv: Task-aware kv cache optimization via semantic
  differentiation of attention heads.
\newblock \emph{arXiv preprint arXiv:2501.15113}, 2025.

\bibitem[Hong et~al.(2024)Hong, Jiang, Qi, Meng, Yu, Zhou, and
  Zhou]{hong2024token}
Xiangyu Hong, Che Jiang, Biqing Qi, Fandong Meng, Mo~Yu, Bowen Zhou, and Jie
  Zhou.
\newblock On the token distance modeling ability of higher rope attention
  dimension.
\newblock \emph{arXiv preprint arXiv:2410.08703}, 2024.

\bibitem[Hsieh et~al.(2024)Hsieh, Sun, Kriman, Acharya, Rekesh, Jia, Zhang, and
  Ginsburg]{hsieh2024ruler}
Cheng-Ping Hsieh, Simeng Sun, Samuel Kriman, Shantanu Acharya, Dima Rekesh, Fei
  Jia, Yang Zhang, and Boris Ginsburg.
\newblock Ruler: What's the real context size of your long-context language
  models?
\newblock \emph{arXiv preprint arXiv:2404.06654}, 2024.

\bibitem[Jiang et~al.(2023)Jiang, Sablayrolles, Mensch, Bamford, Chaplot,
  Casas, Bressand, Lengyel, Lample, Saulnier, et~al.]{jiang2023mistral}
Albert~Q Jiang, Alexandre Sablayrolles, Arthur Mensch, Chris Bamford,
  Devendra~Singh Chaplot, Diego de~las Casas, Florian Bressand, Gianna Lengyel,
  Guillaume Lample, Lucile Saulnier, et~al.
\newblock Mistral 7b.
\newblock \emph{arXiv preprint arXiv:2310.06825}, 2023.

\bibitem[Li et~al.(2024)Li, Huang, Yang, Venkitesh, Locatelli, Ye, Cai, Lewis,
  and Chen]{li2024snapkv}
Yuhong Li, Yingbing Huang, Bowen Yang, Bharat Venkitesh, Acyr Locatelli,
  Hanchen Ye, Tianle Cai, Patrick Lewis, and Deming Chen.
\newblock Snapkv: Llm knows what you are looking for before generation.
\newblock \emph{arXiv preprint arXiv:2404.14469}, 2024.

\bibitem[Liu et~al.(2024)Liu, Feng, Xue, Wang, Wu, Lu, Zhao, Deng, Zhang, Ruan,
  et~al.]{liu2024deepseek}
Aixin Liu, Bei Feng, Bing Xue, Bingxuan Wang, Bochao Wu, Chengda Lu, Chenggang
  Zhao, Chengqi Deng, Chenyu Zhang, Chong Ruan, et~al.
\newblock Deepseek-v3 technical report.
\newblock \emph{arXiv preprint arXiv:2412.19437}, 2024.

\bibitem[Liu et~al.(2023)Liu, Wang, Dao, Zhou, Yuan, Song, Shrivastava, Zhang,
  Tian, Re, et~al.]{liu2023deja}
Zichang Liu, Jue Wang, Tri Dao, Tianyi Zhou, Binhang Yuan, Zhao Song, Anshumali
  Shrivastava, Ce~Zhang, Yuandong Tian, Christopher Re, et~al.
\newblock Deja vu: Contextual sparsity for efficient llms at inference time.
\newblock In \emph{International Conference on Machine Learning}, pages
  22137--22176. PMLR, 2023.

\bibitem[Olsson et~al.(2022)Olsson, Elhage, Nanda, Joseph, DasSarma, Henighan,
  Mann, Askell, Bai, Chen, et~al.]{olsson2022context}
Catherine Olsson, Nelson Elhage, Neel Nanda, Nicholas Joseph, Nova DasSarma,
  Tom Henighan, Ben Mann, Amanda Askell, Yuntao Bai, Anna Chen, et~al.
\newblock In-context learning and induction heads.
\newblock \emph{arXiv preprint arXiv:2209.11895}, 2022.

\bibitem[Oren et~al.(2024)Oren, Hassid, Yarden, Adi, and
  Schwartz]{oren2024transformers}
Matanel Oren, Michael Hassid, Nir Yarden, Yossi Adi, and Roy Schwartz.
\newblock Transformers are multi-state rnns.
\newblock \emph{arXiv preprint arXiv:2401.06104}, 2024.

\bibitem[Peng and Quesnelle(2023)]{peng2023ntk}
Bowen Peng and Jeffrey Quesnelle.
\newblock Ntk-aware scaled rope allows llama models to have extended (8k+)
  context size without any fine-tuning and minimal perplexity degradation,
  2023.

\bibitem[Rajpurkar(2016)]{rajpurkar2016squad}
P~Rajpurkar.
\newblock Squad: 100,000+ questions for machine comprehension of text.
\newblock \emph{arXiv preprint arXiv:1606.05250}, 2016.

\bibitem[Ribar et~al.(2023)Ribar, Chelombiev, Hudlass-Galley, Blake, Luschi,
  and Orr]{ribar2023sparq}
Luka Ribar, Ivan Chelombiev, Luke Hudlass-Galley, Charlie Blake, Carlo Luschi,
  and Douglas Orr.
\newblock Sparq attention: Bandwidth-efficient llm inference.
\newblock \emph{arXiv preprint arXiv:2312.04985}, 2023.

\bibitem[Su et~al.(2024)Su, Ahmed, Lu, Pan, Bo, and Liu]{su2024roformer}
Jianlin Su, Murtadha Ahmed, Yu~Lu, Shengfeng Pan, Wen Bo, and Yunfeng Liu.
\newblock Roformer: Enhanced transformer with rotary position embedding.
\newblock \emph{Neurocomputing}, 568:\penalty0 127063, 2024.

\bibitem[Sun et~al.(2024{\natexlab{a}})Sun, Chang, Bao, Zheng, Zheng, Liu,
  Dong, Chi, and Chen]{sun2024shadowkv}
Hanshi Sun, Li-Wen Chang, Wenlei Bao, Size Zheng, Ningxin Zheng, Xin Liu, Harry
  Dong, Yuejie Chi, and Beidi Chen.
\newblock Shadowkv: Kv cache in shadows for high-throughput long-context llm
  inference.
\newblock \emph{arXiv preprint arXiv:2410.21465}, 2024{\natexlab{a}}.

\bibitem[Sun et~al.(2024{\natexlab{b}})Sun, Chen, Kolter, and
  Liu]{sun2024massive}
Mingjie Sun, Xinlei Chen, J~Zico Kolter, and Zhuang Liu.
\newblock Massive activations in large language models.
\newblock \emph{arXiv preprint arXiv:2402.17762}, 2024{\natexlab{b}}.

\bibitem[Tang et~al.(2024{\natexlab{a}})Tang, Lin, Lin, Han, Hong, Yao, and
  Wang]{tang2024razorattention}
Hanlin Tang, Yang Lin, Jing Lin, Qingsen Han, Shikuan Hong, Yiwu Yao, and
  Gongyi Wang.
\newblock Razorattention: Efficient kv cache compression through retrieval
  heads.
\newblock \emph{arXiv preprint arXiv:2407.15891}, 2024{\natexlab{a}}.

\bibitem[Tang et~al.(2024{\natexlab{b}})Tang, Zhao, Zhu, Xiao, Kasikci, and
  Han]{tang2024quest}
Jiaming Tang, Yilong Zhao, Kan Zhu, Guangxuan Xiao, Baris Kasikci, and Song
  Han.
\newblock Quest: Query-aware sparsity for efficient long-context llm inference.
\newblock \emph{arXiv preprint arXiv:2406.10774}, 2024{\natexlab{b}}.

\bibitem[Team et~al.(2024{\natexlab{a}})Team, Georgiev, Lei, Burnell, Bai,
  Gulati, Tanzer, Vincent, Pan, Wang, et~al.]{team2024gemini}
Gemini Team, Petko Georgiev, Ving~Ian Lei, Ryan Burnell, Libin Bai, Anmol
  Gulati, Garrett Tanzer, Damien Vincent, Zhufeng Pan, Shibo Wang, et~al.
\newblock Gemini 1.5: Unlocking multimodal understanding across millions of
  tokens of context.
\newblock \emph{arXiv preprint arXiv:2403.05530}, 2024{\natexlab{a}}.

\bibitem[Team et~al.(2024{\natexlab{b}})Team, Mesnard, Hardin, Dadashi,
  Bhupatiraju, Pathak, Sifre, Rivi{\`e}re, Kale, Love, et~al.]{team2024gemma}
Gemma Team, Thomas Mesnard, Cassidy Hardin, Robert Dadashi, Surya Bhupatiraju,
  Shreya Pathak, Laurent Sifre, Morgane Rivi{\`e}re, Mihir~Sanjay Kale,
  Juliette Love, et~al.
\newblock Gemma: Open models based on gemini research and technology.
\newblock \emph{arXiv preprint arXiv:2403.08295}, 2024{\natexlab{b}}.

\bibitem[Vaswani(2017)]{vaswani2017attention}
A~Vaswani.
\newblock Attention is all you need.
\newblock \emph{Advances in Neural Information Processing Systems}, 2017.

\bibitem[Wan et~al.(2023)Wan, Wang, Liu, Alam, Zheng, Liu, Qu, Yan, Zhu, Zhang,
  et~al.]{wan2023efficient}
Zhongwei Wan, Xin Wang, Che Liu, Samiul Alam, Yu~Zheng, Jiachen Liu, Zhongnan
  Qu, Shen Yan, Yi~Zhu, Quanlu Zhang, et~al.
\newblock Efficient large language models: A survey.
\newblock \emph{arXiv preprint arXiv:2312.03863}, 2023.

\bibitem[Wu et~al.(2024)Wu, Wang, Xiao, Peng, and Fu]{wu2024retrieval}
Wenhao Wu, Yizhong Wang, Guangxuan Xiao, Hao Peng, and Yao Fu.
\newblock Retrieval head mechanistically explains long-context factuality.
\newblock \emph{arXiv preprint arXiv:2404.15574}, 2024.

\bibitem[Xiao et~al.(2023)Xiao, Tian, Chen, Han, and Lewis]{xiao2023efficient}
Guangxuan Xiao, Yuandong Tian, Beidi Chen, Song Han, and Mike Lewis.
\newblock Efficient streaming language models with attention sinks.
\newblock \emph{arXiv preprint arXiv:2309.17453}, 2023.

\bibitem[Xiao et~al.(2024)Xiao, Tang, Zuo, Guo, Yang, Tang, Fu, and
  Han]{xiao2024duoattention}
Guangxuan Xiao, Jiaming Tang, Jingwei Zuo, Junxian Guo, Shang Yang, Haotian
  Tang, Yao Fu, and Song Han.
\newblock Duoattention: Efficient long-context llm inference with retrieval and
  streaming heads.
\newblock \emph{arXiv preprint arXiv:2410.10819}, 2024.

\bibitem[Zhang et~al.(2023)Zhang, Sheng, Zhou, Chen, Zheng, Cai, Song, Tian,
  R{\'e}, Barrett, et~al.]{zhang2023h2o}
Zhenyu Zhang, Ying Sheng, Tianyi Zhou, Tianlong Chen, Lianmin Zheng, Ruisi Cai,
  Zhao Song, Yuandong Tian, Christopher R{\'e}, Clark Barrett, et~al.
\newblock H2o: Heavy-hitter oracle for efficient generative inference of large
  language models.
\newblock \emph{Advances in Neural Information Processing Systems},
  36:\penalty0 34661--34710, 2023.

\bibitem[Zheng et~al.(2024)Zheng, Wang, Huang, Song, Yang, Tang, Xiong, and
  Li]{zheng2024attention}
Zifan Zheng, Yezhaohui Wang, Yuxin Huang, Shichao Song, Mingchuan Yang,
  Bo~Tang, Feiyu Xiong, and Zhiyu Li.
\newblock Attention heads of large language models: A survey.
\newblock \emph{arXiv preprint arXiv:2409.03752}, 2024.

\end{thebibliography}
\bibliographystyle{assets/plainnat}

\appendix
\onecolumn

\section{Additional Experimental Details}
\label{sec:apx-exp-details}
In this section we present additional details for the experiments.

\begin{figure*}[t]
    \centering    
        \centering
            \includegraphics[width=\linewidth]{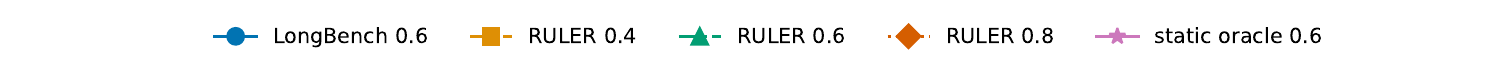}

                \begin{subfigure}[b]{0.47\linewidth}
\includegraphics[width=\linewidth]{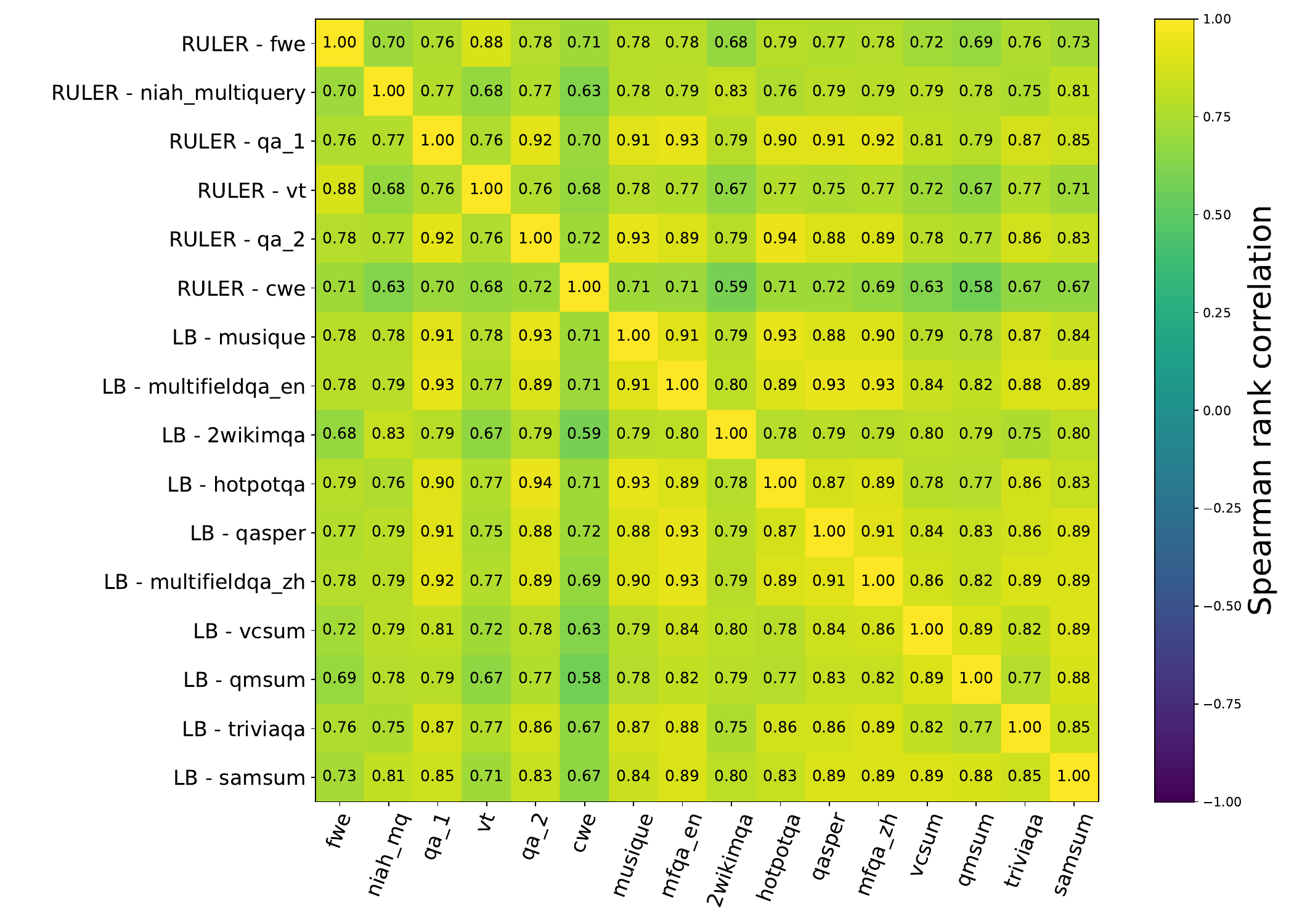}

        \caption{ Spearman rank correlation of heads}
        \end{subfigure}
                \begin{subfigure}[b]{0.52\linewidth}
\includegraphics[width=\linewidth]{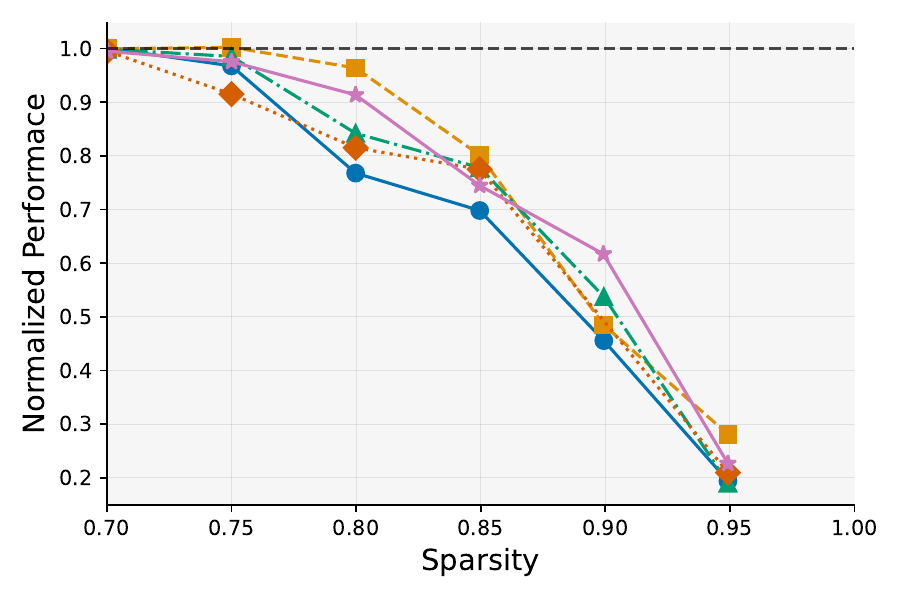}
    
        \caption{ Ablation over datasets for static criterion}
        \label{fig:static}
        \end{subfigure}
    \caption{\small \textbf{a)}  The Spearman rank correlation of the attention heads ordered by the fraction of times labeled as Local Heads by the oracle criterion with $\tau=0.6$. We see a high correlation among all tasks. b) Ablations for the static criterion using different datasets (LongBench, RULER and specific RULER task, called oracle) and threshold $\thrsstatic$ to label the heads. We use Llama3-8B on RULER 8k.} 
    \label{fig:staticablations}
\end{figure*}

\paragraph{Additional details for the methods }
The best way to select the ``right'' subset of attention heads for the static criterion is still widely understudied. In particular, it poses  the fundamental challenge of which dataset should be chosen to select the heads in advance. Since we are primarily interested in how much query-adaptivity helps to improve, we compare against a \textbf{static oracle} criterion, that uses the prompts for evaluation to decide which heads are sued as static heads. Moreover, we also implement \textbf{static RULER}, using the prompts from the RULER task. We present additional ablations for the choice of the static criterion in Figure~\ref{fig:staticablations}.
Similar to \citet{wu2024retrieval,tang2024razorattention}, we measure head patterns in a synthetic retrieval task, and select heads via the following  simple \textbf{static criterion}: 
\begin{itemize}
    \item \textit{Step 1}: Generate responses for selected prompts using full attention (for LongBench, GSM8k and MBPP tasks) or the approximate attention from the oracle criterion with $\thrsoracle =0.6$ (RULER tasks). Compute the percentage of times each head is labeled as local window by the oracle criterion from Equation~\eqref{eq:oracle} with threshold $\thrsstatic$.
\item
\textit{Step 2}: Calculate the $(1-\alpha)$-quantile of these percentages across all heads $h$. Label heads below the threshold as \textit{long-context} ($c^h_{\text{static}} = 0$) and those above as \textit{local} ($c^h_{\text{static}} = 1$). These labels are query-independent.
\end{itemize}

We further refer the reader to Appendix~\ref{sec:keys} for how we compute the moments used by  \textbf{QAdA}, for which we devote an entire section. 

\paragraph{Choices for thresholds} We ablate over the various thresholds $\thrsoracle, \thrsapprox \in $ (0.1, 0.2, 0.3, 0.4, 0.5, 0.6, 0.7, 0.8, 0.9, 0.95, 0.99, 0.995), as well as 
$\alpha \in $ (0.05, 0.1, 0.15, 0.2, 0.25, 0.3, 0.35, 0.4, 0.45, 0.5, 0.55, 0.6) with $\thrsstatic=0.6$. We ran additional ablations in Figure~\ref{fig:static} for $\thrsstatic$ confirming that the choice $\thrsstatic=0.6$ yields robust performance across all tasks.

\paragraph{RULER tasks} The RULER benchmark \citep{hsieh2024ruler} consists of a collection of synthetic tasks with varying prompt sizes. These tasks are designed to challenge the model's capabilities in processing long-context information.
We choose the two Q/A tasks, ``qa-1'' and ``qa-2'', the two aggregation tasks: common words extraction ``cwe'' and frequent words extraction ``fwe'', the variable tracing task ``vt'', and the multiquery needle-in-a-haystack task ``niah''. Especially, the two aggregation tasks ``fwe'' and ``cwe'' are known to be difficult baselines for achieving accuracy using efficient sparse attention mechanisms (see the discussion in \citet{chen2024magicpig}).

\paragraph{LongBench tasks} The LongBench benchmark contains a selection of challenging real-world and synthetic tasks, including single-doc QA, multi-doc QA, summarization, and few-shot learning.
We use a selection of tasks from the LongBench dataset for which the standard model achieves at least decent scores. We evaluate on the tasks: (Single-Document QA): ``qasper'', ``multifieldqa-en'', ``multifieldqa-zh'', ``narrativeqa''; (Multi-Document QA): ``2wikimqa'', ``musique'', ``hotpotqa''; (Summarization): ``qmsum'', ``vcsum''; and (Few-shot Learning): ``triviaqa''.

\paragraph{Long-context GSM8k and MBPP datasets}

In addition to the two standard benchmarks, RULER and LongBench, we also construct our own long-context tasks based on the reasoning task GSM8k \citep{cobbe2021training} and the code-generation task MBPP \citep{austin2021program}. We use the standard evaluation protocol, but instead of using only the ``correct'' few-shot examples, we select 55 few-shot examples in the same format generated from the SQUAD \citep{rajpurkar2016squad} dataset, as well as 5 actual few-shot examples (highlighted in green). We provide fragments of the example prompts below. The resulting context lengths are $\approx 10k$ for GSM8k and $\approx 11k$ for MBPP.

For these two tasks, we always use the pre-trained Llama3-8B parameter model \citep{dubey2024llama}, instead of the instruction fine-tuned variant. The reason for choosing the pre-trained model is that the instruction fine-tuned model can solve these tasks without the need for few-shot examples, while the pre-trained model crucially depends on few-shot examples. Since these examples are hidden in a long context, the task becomes challenging, and the model requires retrieving information from tokens far away in order to achieve high accuracy on the task.

\section{Computing the moment statistics}
\label{sec:keys}
We discuss in this section more formally how we obtain the moment statistics as sketched in Section~\ref{sec:moments}.

\begin{figure*}
    \centering    

\begin{subfigure}[b]{0.24\linewidth}
        \centering
        \includegraphics[width=\linewidth]{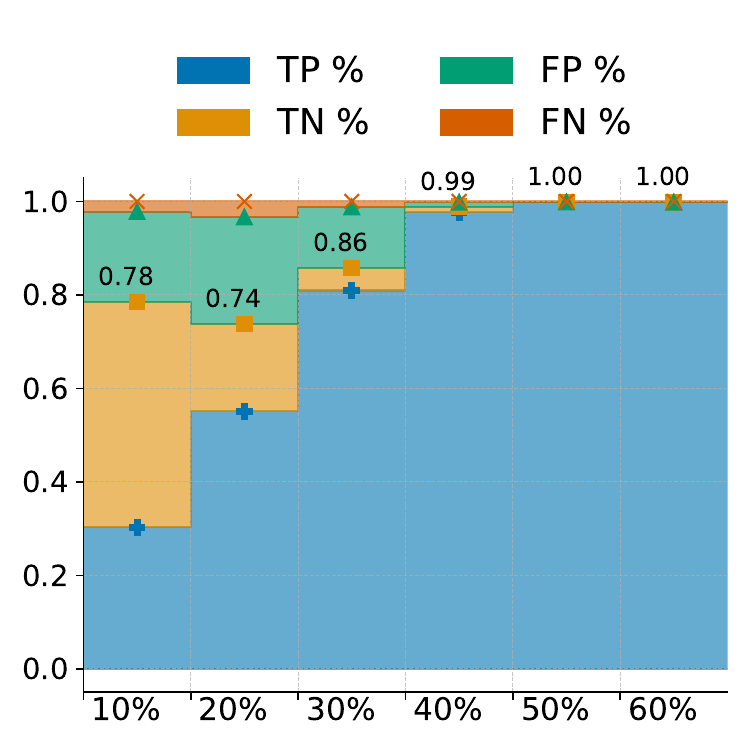}
        \caption{oracle vs adaptive}
        \label{fig:accuracy}
    \end{subfigure}
        \begin{subfigure}[b]{0.24\linewidth}
        \centering
        \includegraphics[width=\linewidth]{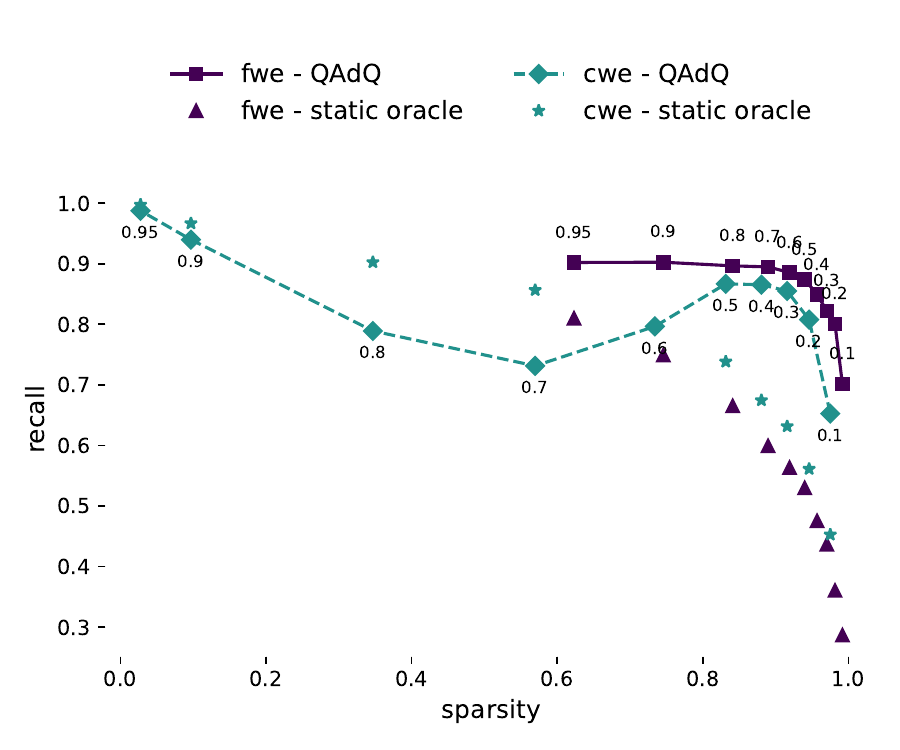}
        \caption{recall of aggregation}
        \label{fig:recalla}
    \end{subfigure}
        \begin{subfigure}[b]{0.24\linewidth}
        \centering
        \includegraphics[width=\linewidth]{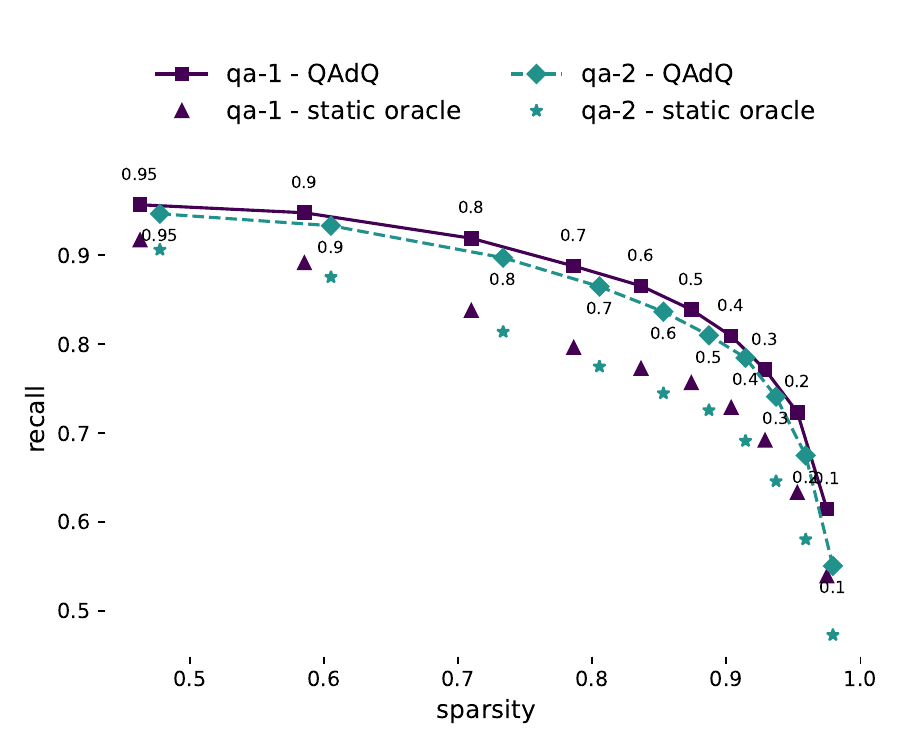}
        \caption{recall of Q/A}
        \label{fig:recallb}
    \end{subfigure}
        \begin{subfigure}[b]{0.24\linewidth}
        \centering
        \includegraphics[width=\linewidth]{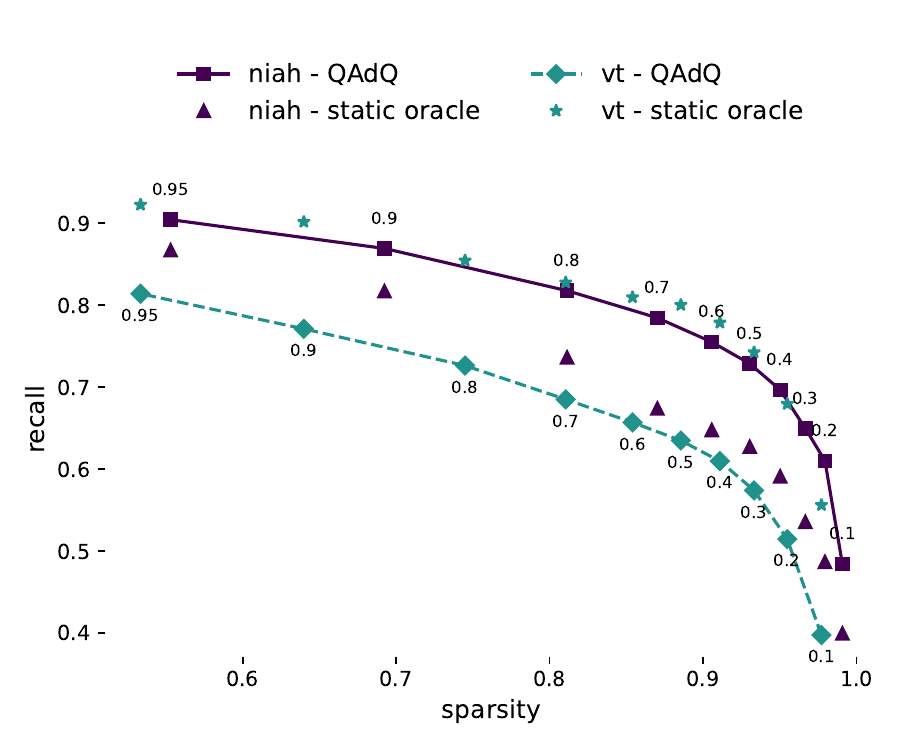}
        \caption{recall of retrieval}
        \label{fig:recallc}
    \end{subfigure}

    \caption{\small \textbf{a)}  Accuracy and fraction of true/false  positives/negatives for the 10\% quantiles of the heads (labeled as local heads) for the adaptive criterion with  $\thrsoracle=\thrsapprox=0.6$ on the  RULER benchmark with sequence length 8k. 
    \textbf{b,c,d)} The recall values of long-context heads selected by the oracle criterion for various thresholds $\thrsoracle$ when using the static and adaptive oracle criteria as a function of the average sparsity (percentage of local heads). We adjust the thresholds $\alpha$ (with $\thrsstatic = \thrsoracle$) and $\thrsapprox$ to achieve matching sparsity levels. Annotations indicate the specific oracle thresholds $\thrsoracle$.  We use Llama3-8B on RULER 8k.} 
    \vspace{-0.2in}
\end{figure*}

\paragraph{Option 1 (current prompt):} In this case, after pre-filling, we compute the moment statistics for each head as described in Section~\ref{sec:moments}. Note that for grouped-query attention \citep{ainslie2023gqa}, as used by Llama, we naturally use the same moments for each query in the group since these heads share the same keys. During generation, we keep the moment statistics fixed and do not update them after predicting each token. This is because we always generate sequences of length less than $256$, so updating the statistics has only a limited influence. However, when generating long sequences consisting of thousands of tokens, we would expect that updating the moments during generation becomes beneficial for performance.

\paragraph{Option 2 (other prompt):} In this case, we perform a single forward pass using one of the three choices as prompts: \textit{random word prompt}, which simply permutes words from a Wikipedia article (including the HTML syntax); \textit{wiki prompt}, where we concatenate Wikipedia articles; and \textit{single words prompt}, where we repeat the word "observation." As we showed in Section~\ref{sec:ablations}, the content of the prompt is not important as long as there is enough "diversity." However, we found that the length of the sequence is crucial. Therefore, we store all keys from the forward pass of this prompt. During generation, when predicting the next tokens for a given prompt, we load the keys from the specific \textit{other prompt} and generate the moments using the first $T-1024$ keys, where $T$ is the sequence length of the current prompt. The reason for choosing minus $1024$ is because, as we saw in Figure~\ref{fig:seqlen_prompt}, the performance is robust to keys generated from shorter prompts than the actual sequence but suffers significantly in performance for longer ones. As an alternative implementation, one could also pre-compute the moments for lengths of fixed intervals and load the corresponding moment after pre-filling before starting the generation.

\section{Recall of Attention Heads}
\label{sec:recall}
In this section, we analyze how well our adaptive criterion from Section~\ref{sec:method} can recall the heads selected by the oracle criterion; in other words, how effectively it serves as a proxy for the oracle. We always use the current prompt (Option 1) to generate the moment statistics.

\paragraph{Accuracy}

We generate responses using standard dense attention and store the scores used to compare the two criteria using the current prompt to generate the moments. For each task, we group the heads into $10\%$ quantiles based on the percentage of times the oracle criterion has been satisfied. For each quantile (averaged over the six selected RULER tasks), we show the fraction of true positives, true negatives, false positives, and false negatives, where a true positive means that both the oracle and adaptive criteria labeled a head as a local head.

We find that the adaptive criterion always correctly identifies the top $50\%$ of the heads that are consistently local heads. Moreover, we find even higher accuracies for the lower quantiles where heads vary between local and long-context. Interestingly, we see that the false negative rate is much lower than the false positive rate for these heads. As a result, the adaptive criterion selects fewer heads than the oracle criterion. This observation is counter-intuitive to the observations made in Section~\ref{sec:rec}, where we observed that our adaptive criterion tends to select more heads than the oracle criterion for the same threshold. The explanation here is that in this section we compare the criterion on scores obtained when using standard full attention. This is necessary to allow a direct comparison between the two criteria. In contrast, in Section~\ref{sec:rec} we compare the average sparsity when using the approximate attention that approximates all labeled heads by a local window.

\paragraph{Recall of long-context heads.} We further compare our adaptive criterion  with the oracle  static criterion in their ability to identify long-context heads selected by the oracle criterion. 
We show in Figure~\ref{fig:recalla}-\ref{fig:recallc} the recall value of long-context heads selected by the oracle criterion for different oracle thresholds $\thrsoracle$ as a function of the sparsity (fraction of heads labeled as local heads by the oracle criterion). 
To allow for a direct comparison between static and adaptive, we  choose $\thrsapprox$, resp. quantile $\alpha$ (with $\thrsstatic = \thrsoracle$), such that the average sparsity is the same as the one of the oracle criterion. We plot the curves for all (selected)  RULER tasks, and find that our test achieves consistently a higher recall value than the oracle static assignment (except for the ``vt'' task, for which the \textit{current prompt} choice for the moments breaks down, as discussed in Section~\ref{sec:ablations}).

\begin{table}[t]
\centering
\begin{tabular}{@{}lccc@{}}
\toprule
Method & all & top 20\% & top 10\% \\
& $\mu \pm \sigma$ & $\mu \pm \sigma$ & $\mu \pm \sigma$ \\
\midrule\midrule
 & \multicolumn{3}{c}{RULER 8k task ``fwe''} \\ 
\midrule\midrule
Log error & $0.41 \pm 0.58$ & $0.50 \pm 0.98$ & $0.57 \pm 1.27$ \\
Dist. local & $3.44 \pm 1.73$ & $1.78 \pm 1.38$ & $1.54 \pm 1.23$ \\
Gaussian opt. & $0.15 \pm 0.18$ & $0.14 \pm 0.21$ & $0.15 \pm 0.25$ \\
\midrule\midrule
 & \multicolumn{3}{c}{RULER 8k task ``Q/A-2''} \\
\midrule\midrule
Log error & $0.37 \pm 0.52$ & $0.63 \pm 0.75$ & $0.74 \pm 0.83$ \\
Dist. local & $2.80 \pm 1.55$ & $1.17 \pm 0.98$ & $1.29 \pm 1.08$ \\
Gaussian opt.  & $0.18 \pm 0.22$ & $0.25 \pm 0.34$ & $0.29 \pm 0.40$ \\
\bottomrule
\end{tabular}
\caption{The mean and standard deviation for the terms  log difference  $|\log A^{\text{bulk}} - (\log(T^{\text{bulk}}) + \mu_s + \sigma_s^2/2)|$ (Log error) and $|\log A^{\text{bulk}} - \log A^{\text{local}}|$ (Dist. local) for all heads (first column) and the 20\% and 10\% percentiles of heads most often labeled as local heads by the oracle criterion with $\thrsoracle=0.6$. We further show the ``Log error'' when replacing the scores by i.i.d.~Gaussian samples instead with matching mean and variance. This indicates the achievable error assuming that the Gaussian approximation holds true.  We use Llama3-8B on RULER 8k.}\label{tab:comparison}
\vspace{-0.1in}
\end{table}

\section{Discussion: Gaussian Approximation}

 \label{apx:gaussian}
In this section, we further discuss the Gaussian approximation exploited  by our criterion in Section~\ref{sec:method}. We divide the discussion into multiple paragraphs.  

\paragraph{Approximatin error} We wonder what is the approximation error arising from Equation~\eqref{eq:gaussianapprox}. 
We show in Table~\ref{tab:comparison}  the average log difference  $|\log A^{\text{bulk}} - (\log(T^{\text{bulk}}) + \mu_s + \sigma_s^2/2)|$  (first row)  between the un-normalized mass of the bulk and our Gaussian approximation from Equation~\eqref{eq:gaussianapprox}.  Taking the exponent, we find that the Gaussian approximation is typically off by a factor of $\approx 2-5$, and thus clearly imprecise. In comparison, in the third row, we show the same statistics, when replacing the scores by i.i.d~samples from a Gaussian distribution with matching mean and variance. This error captures the ``optimal'' error given that Gaussian actually holds. As we can see, this error is significantly smaller. 

Nevertheless, we are effectively interested in whether the Gaussian assumption suffices to make an accurate prediction on whether the head is a local or long-context head. To that end, we also compare in the second row the average log difference  $|\log A^{\text{bulk}} -  \log A^{\text{local}}|$. Indeed, if this distance is much larger than the average log error arising form the Gaussian approximation, we expect our criterion to nevertheless be accurate. As we observe, this is the case. Taking again the exponent, we  find that the $A^{\text{bulk}} $ and $A^{\text{local}}$ typically differ by  factors around $\approx 15-50$. Interestingly, however, we see that the gap becomes more narrow when only considering the top 20\% (resp. 10\%) of heads most frequently selected by the oracle criterion as long-context heads. Finally, we also show the average standard deviation.

\section{Additional Experiments}

\label{sec:additional_exps}

\begin{figure*}[t]
    \centering    
        \centering
        \includegraphics[width=0.8\linewidth]{plots/ablations/mean_legend.pdf}

        \begin{subfigure}[b]{0.24\linewidth}
        \includegraphics[width=\linewidth]{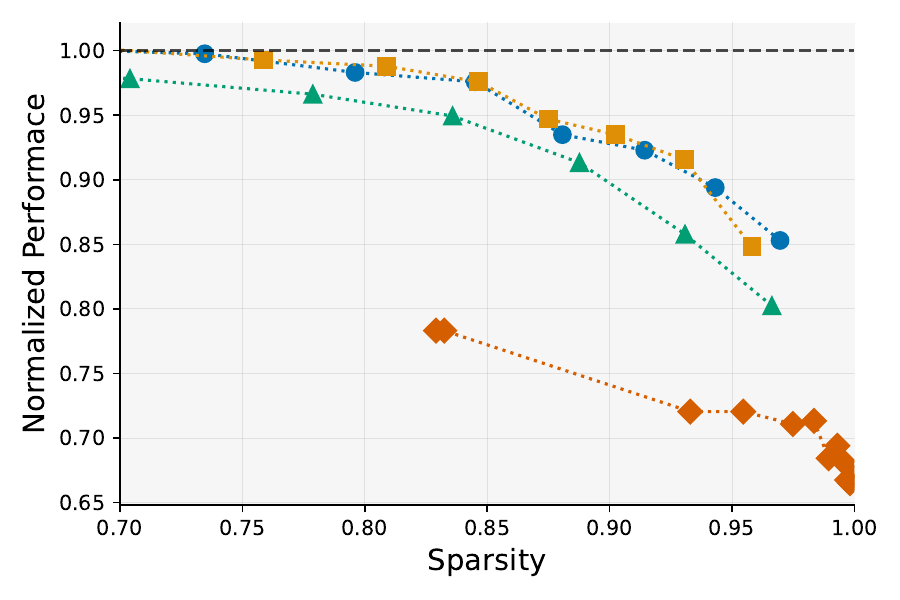}
                            
        \caption{ ``qa-1'' task}
        \end{subfigure}
        \begin{subfigure}[b]{0.24\linewidth}
        \includegraphics[width=\linewidth]{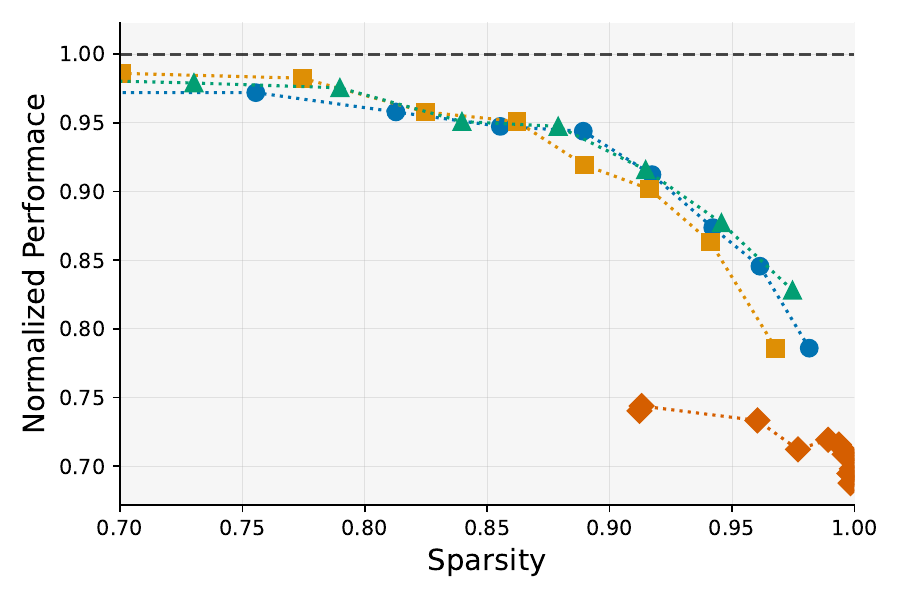}
                            
        \caption{ ``qa-2'' task}
        \end{subfigure}
            \begin{subfigure}[b]{0.24\linewidth}
        \includegraphics[width=\linewidth]{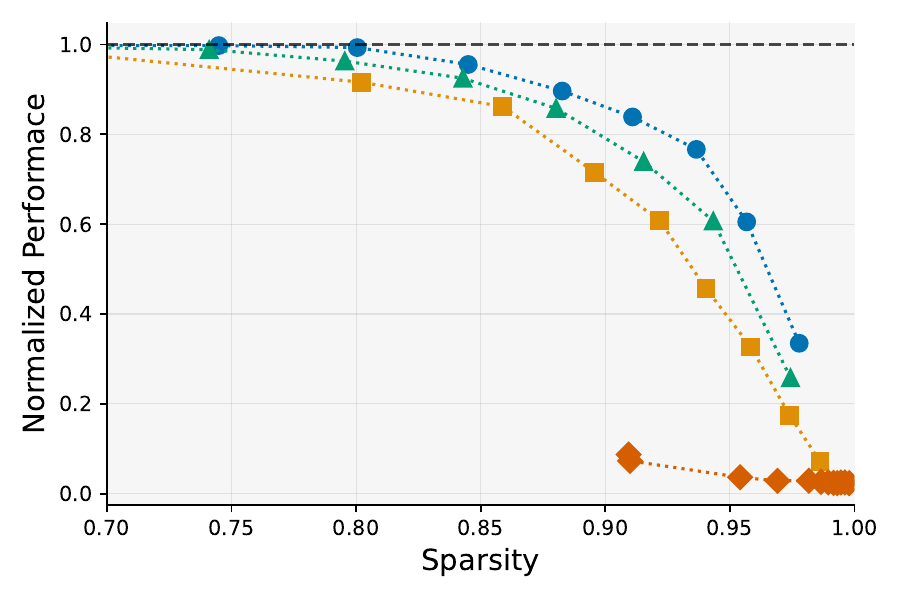}
                            
        \caption{ ``niah'' task}
        \end{subfigure}
            \begin{subfigure}[b]{0.24\linewidth}
        \includegraphics[width=\linewidth]{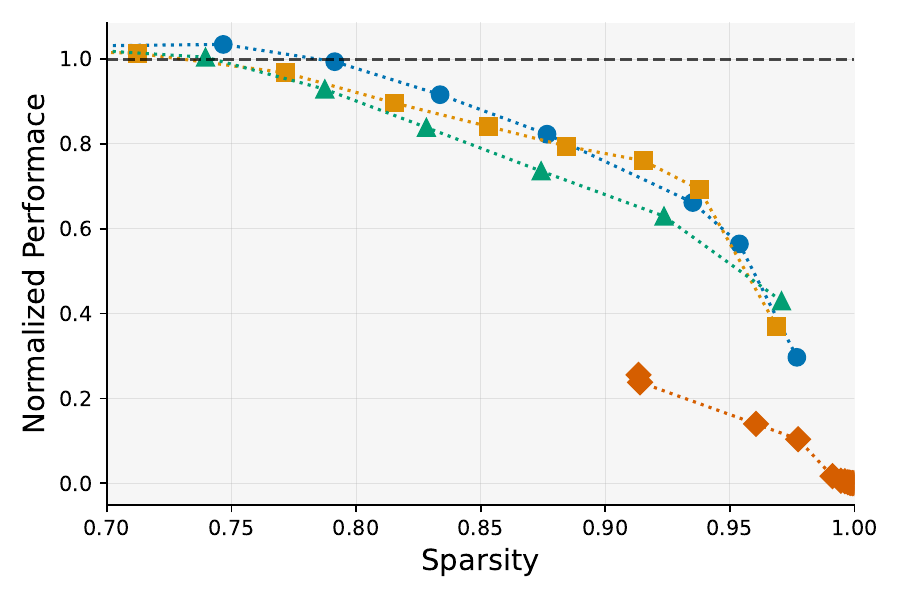}
                            
        \caption{ ``cwe'' task}
        \end{subfigure}

    \caption{\small  Ablations for varying prompts. Same as Figure~\ref{fig:prompts} and \ref{fig:promptsfwe} for the additional RULER $8$k tasks using Llama 3-8B.} 
    \label{fig:ablations-vt-extra}
\end{figure*}

\paragraph{Ablations for the choice of the prompts}
We show in Figure~\ref{fig:ablations-vt-extra} the plots for the other RULER tasks for the ablations for the choice of the prompt in Figures~\ref{fig:prompts},\ref{fig:promptsfwe} in Section~\ref{sec:ablations}.

\begin{figure}
    \centering
    \includegraphics[width=\linewidth]{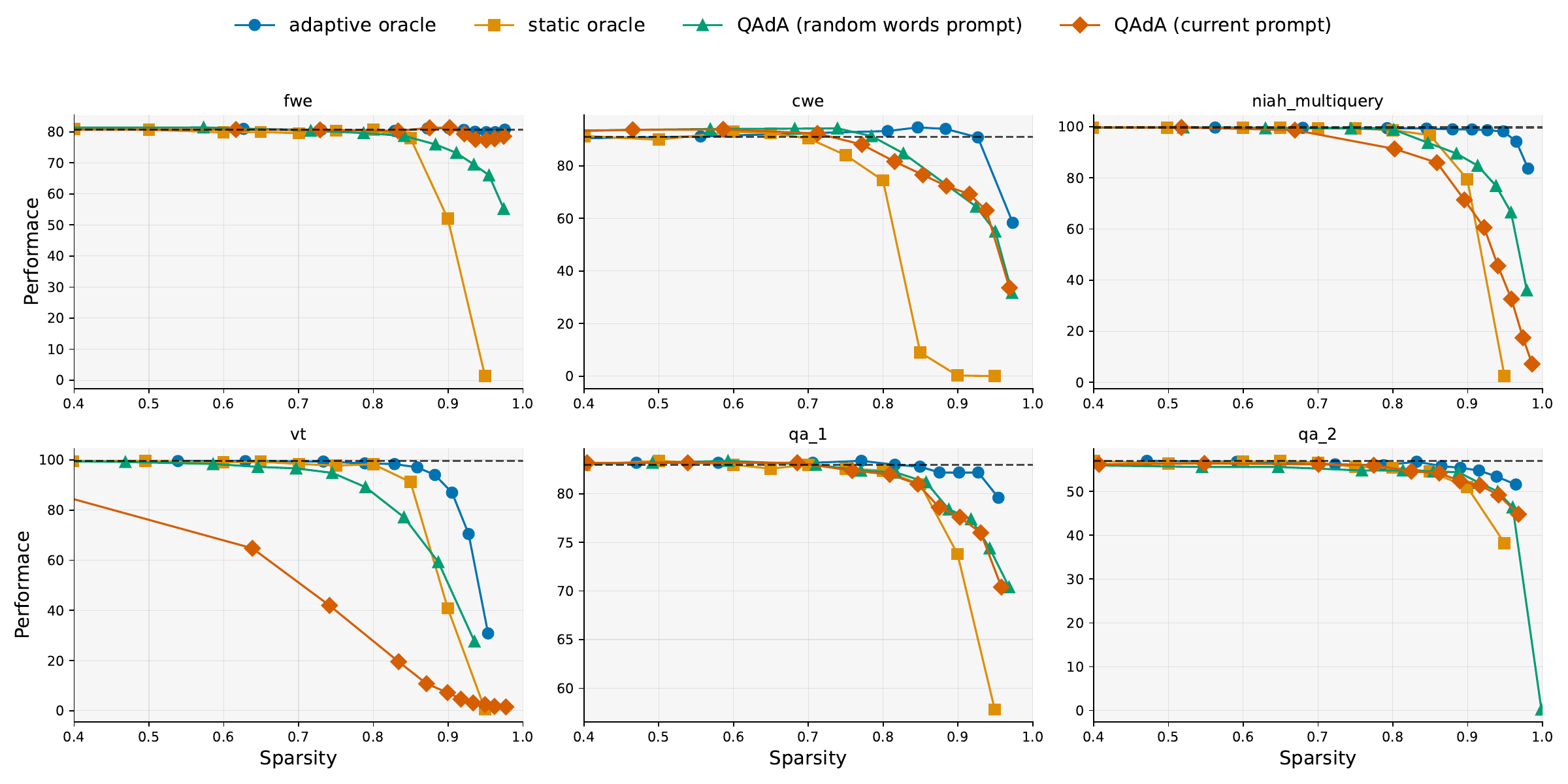}
    \caption{ Performances for individual tasks for RULER $8$k using Llama-3 8B as in Figure~\ref{fig:compare-approx}}
    \label{fig:llama8k}
\end{figure}

\begin{figure}
    \centering
    \includegraphics[width=\linewidth]{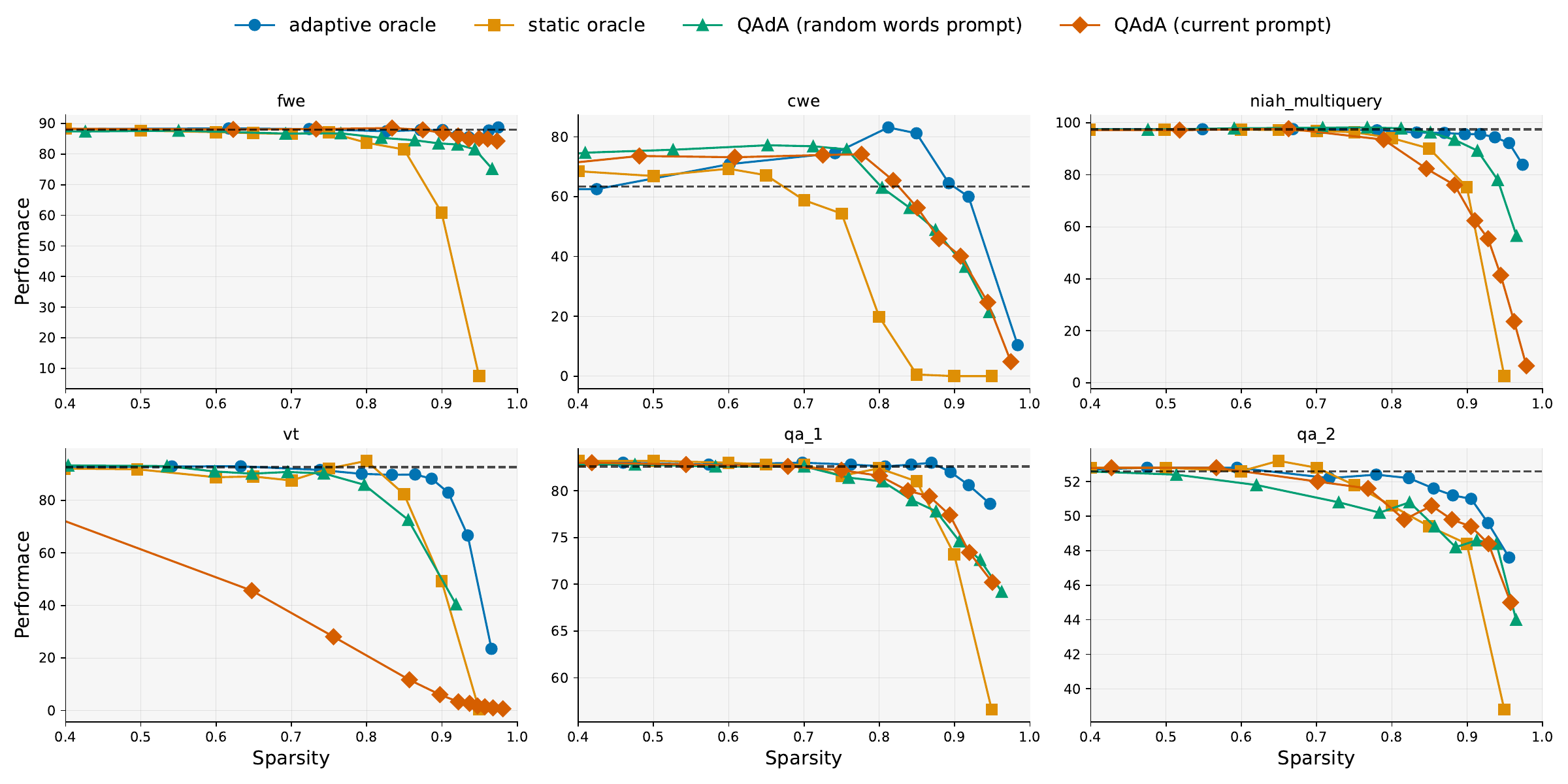}
    \caption{ Performances for individual tasks for RULER $16$k using Llama-3 8B as in Figure~\ref{fig:compare-approx}}
    \label{fig:llama16k}
\end{figure}

\begin{figure}
    \centering
    \includegraphics[width=\linewidth]{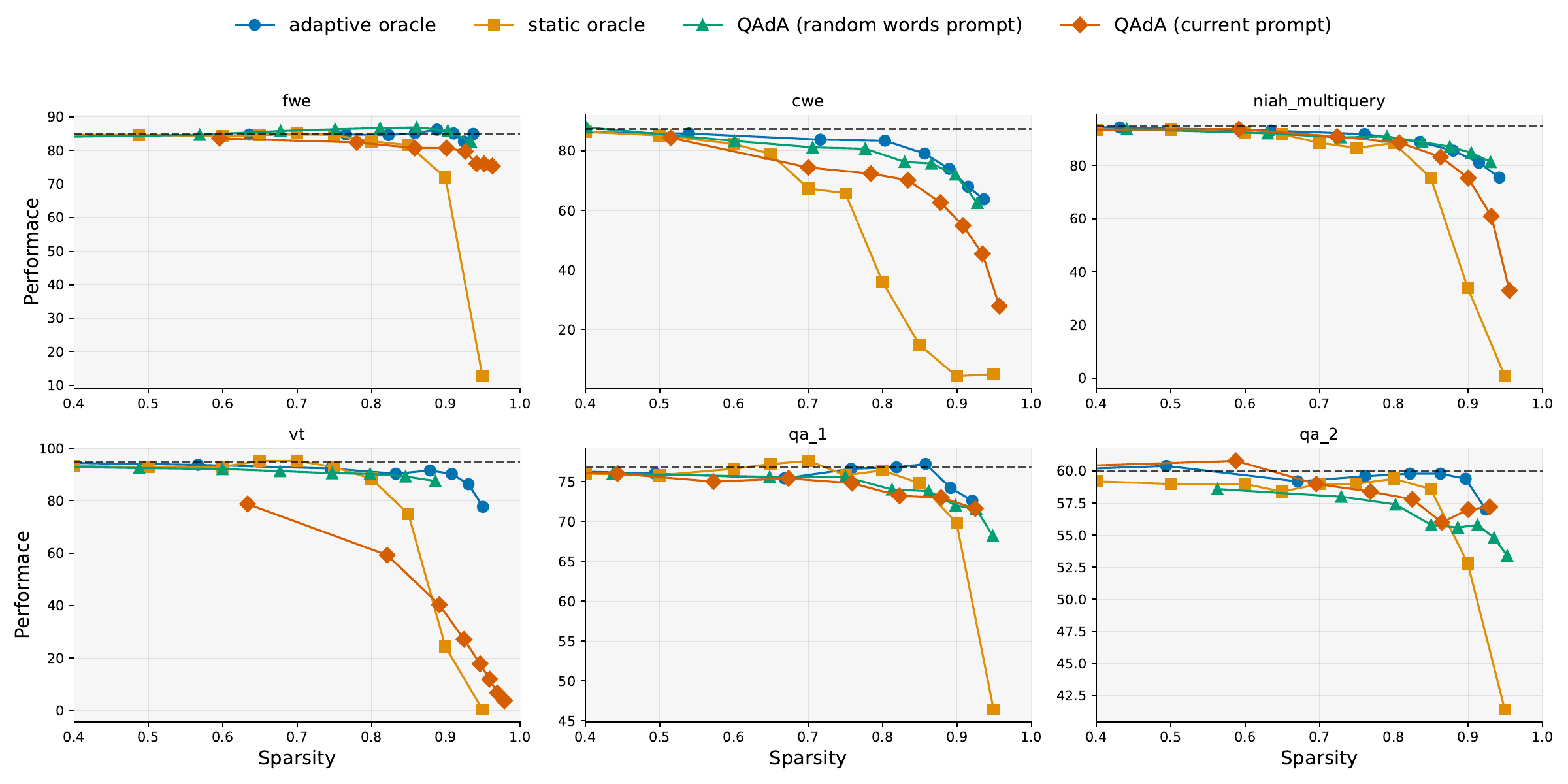}
    \caption{ Performances for individual tasks for RULER $8$k using Mistral-7B as in Figure~\ref{fig:compare-approx}}
    \label{fig:mistral8k}
\end{figure}

\begin{figure}
    \centering
    \includegraphics[width=\linewidth]{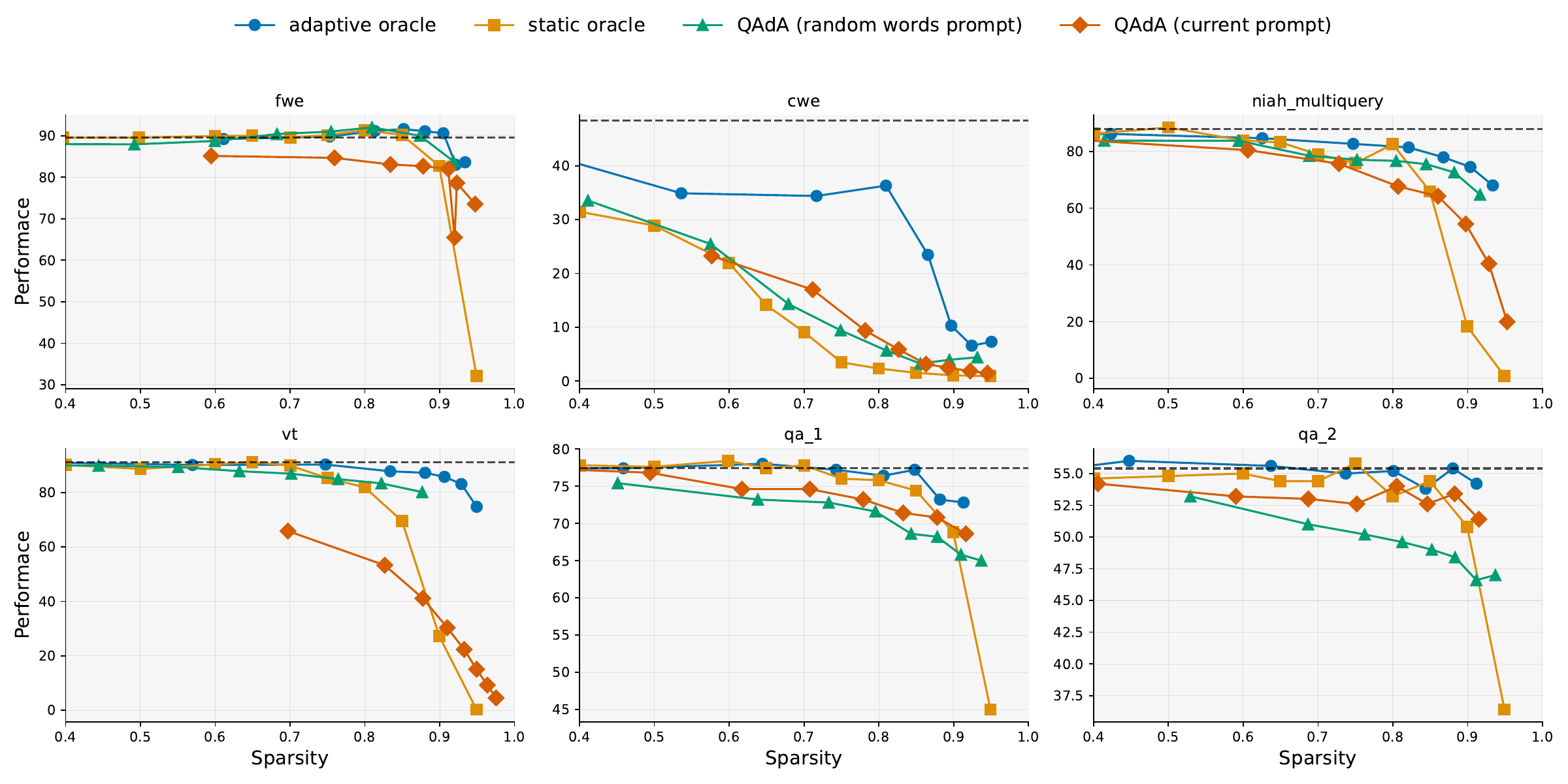}
    \caption{ Performances for individual tasks for RULER $16$k using Mistral-7B as in Figure~\ref{fig:compare-approx}}
    \label{fig:mistral16k}
\end{figure}

\begin{figure}
    \centering
    \includegraphics[width=\linewidth]{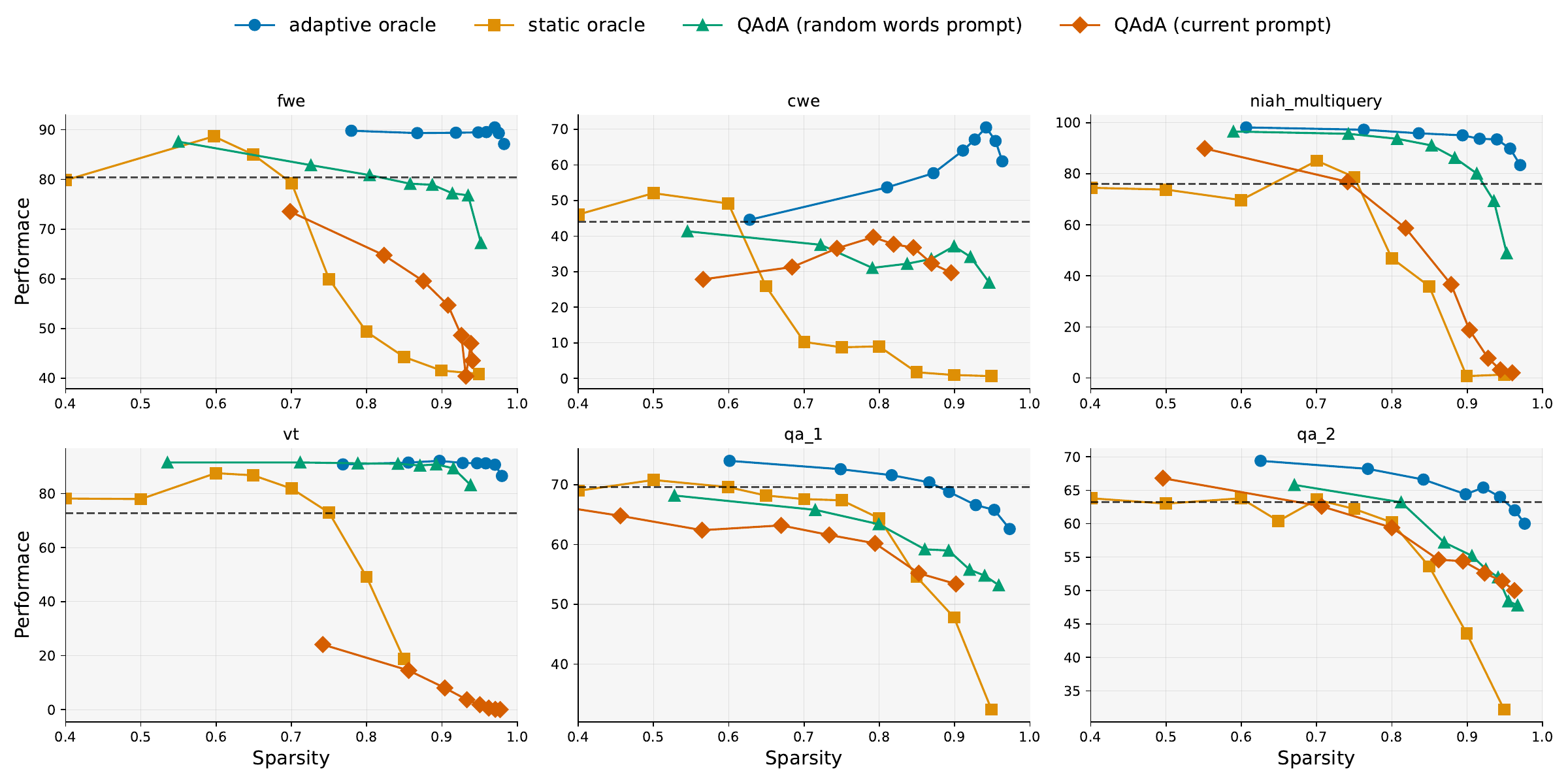}
    \caption{ Performances for individual tasks for RULER $8$k using Qwen-7B as in Figure~\ref{fig:compare-approx}}
    \label{fig:qwen9k}
\end{figure}

\begin{figure}
    \centering
    \includegraphics[width=\linewidth]{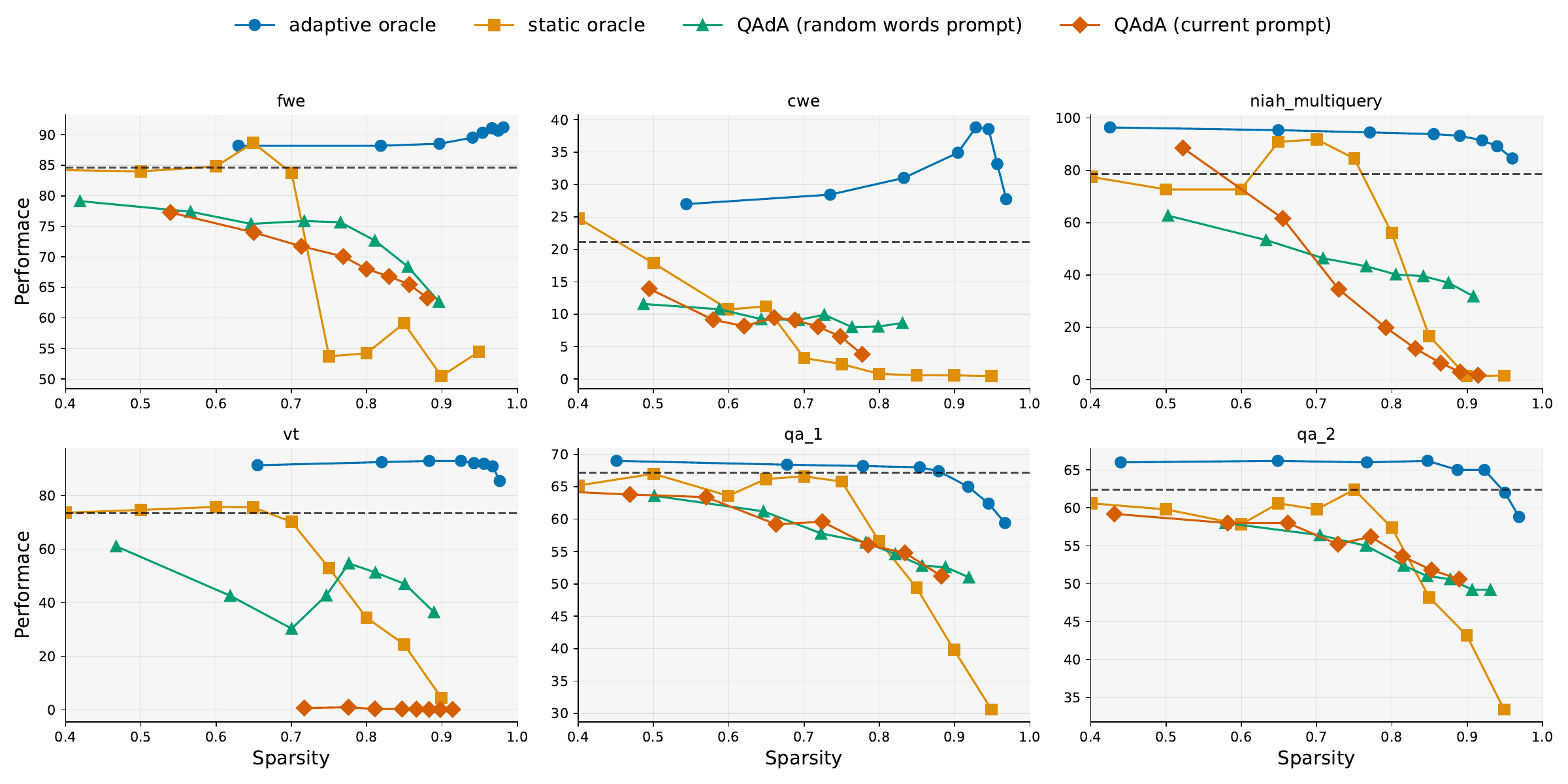}
    \caption{ Performances for individual tasks for RULER $16$k using Qwen-7B as in Figure~\ref{fig:compare-approx}}
    \label{fig:qwen16k}
\end{figure}

\begin{figure}
    \centering
    \includegraphics[width=\linewidth]{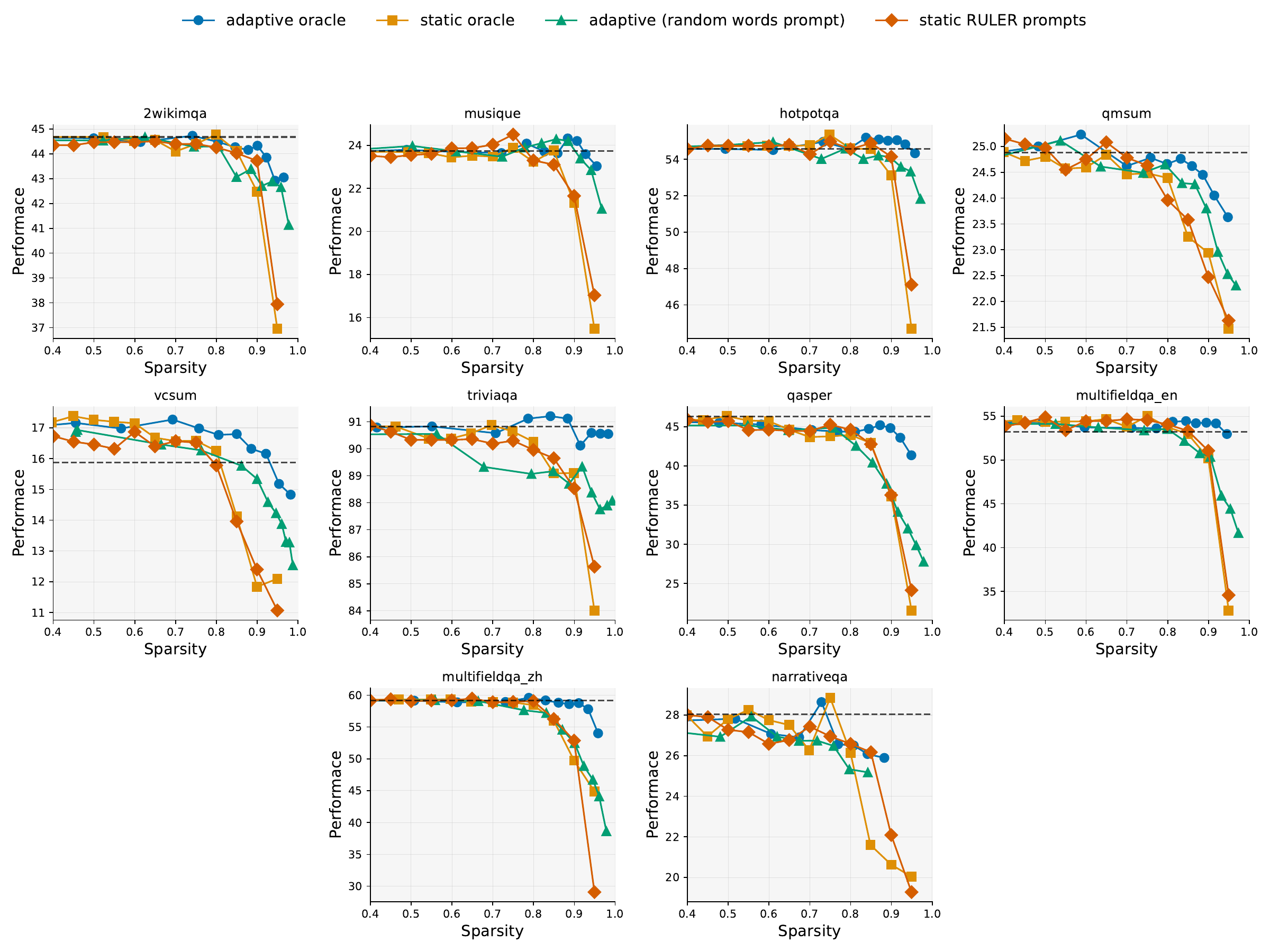}
    \caption{\small Performances for individual tasks for LongBench as in Figure~\ref{fig:longbench}}
    \label{fig:longappendix}
\end{figure}

\paragraph{Performances for individual tasks}
We showed in Figures~\ref{fig:compare-approx} and \ref{fig:longbench} the aggregated performances over the tasks. For completeness, we further show in Figures~\ref{fig:llama8k}-\ref{fig:longappendix} the performances for the individual tasks. We further also show the performance of QAdA (current prompt). Interestingly, we observe that the using the random words prompt (Option 2) for generating the keys overwhelmingly often outperforms the use of the current prompt (Option 1). We leave an explanation for this intriguing finding as a task for future work.

\begin{figure*}
\begin{tcolorbox}[
    title=Example Prompt for long-context MBPP,
    width=\textwidth,
    colback=white,
    left=5pt,
    right=5pt,
    top=5pt,
    bottom=5pt
]
[...]\\
Q: Due to extreme variation in elevation, great variation occurs in the climatic conditions of Himachal . The climate varies from hot and subhumid tropical in the southern tracts to, with more elevation, cold, alpine, and glacial in the northern and eastern mountain ranges. The state has areas like Dharamsala that receive very heavy rainfall, as well as those like Lahaul and Spiti that are cold and almost rainless. Broadly, Himachal experiences three seasons: summer, winter, and rainy season. Summer lasts from mid-April till the end of June and most parts become very hot (except in the alpine zone which experiences a mild summer) with the average temperature ranging from 28 to 32 °C (82 to 90 °F). Winter lasts from late November till mid March. Snowfall is common in alpine tracts (generally above 2,200 metres (7,218 ft) i.e. in the higher and trans-Himalayan region).\\
What is the climate like?\\
A: varies from hot and subhumid tropical  The answer is varies from hot and subhumid tropical.\\\\
\textcolor{darkgreen}{
Q: James decides to buy a new bed and bed frame.  The bed frame is \$75 and the bed is 10 times that price.  He gets a deal for 20\% off.  How much does he pay for everything?\\
A: The bed cost 75*10=\$750\\
So everything cost 750+75=\$825\\
He gets 825*.2=\$165 off\\
So that means he pays 825-165=\$660 The answer is 660.}
\\\\
\textcolor{darkgreen}{
Q: Liz sold her car at 80\% of what she originally paid. She uses the proceeds of that sale and needs only \$4,000 to buy herself a new \$30,000 car. How much cheaper is her new car versus what she originally paid for her old one?\\
A: If Liz needs only \$4,000 to buy a new \$30,000 car, that means she has \$30,000-\$4,000=\$26,000 from the proceeds of selling her old car\\
If she sold her car at 80\% of what she originally paid for and sold it for \$26,000 then she originally paid \$26,000/80\% = \$32,500 for her old car\\
If she paid \$32,500 for her old car and the new one is \$30,000 then, the new one is \$32,500-\$30,000 = \$2,500 cheaper The answer is 2500.}\\\\
Q: Unlike in multicellular organisms, increases in cell size (cell growth) and reproduction by cell division are tightly linked in unicellular organisms. Bacteria grow to a fixed size and then reproduce through binary fission, a form of asexual reproduction. Under optimal conditions, bacteria can grow and divide extremely rapidly, and bacterial populations can double as quickly as every 9.8 minutes. In cell division, two identical clone daughter cells are produced. Some bacteria, while still reproducing asexually, form more complex reproductive structures that help disperse the newly formed daughter cells. Examples include fruiting body formation by Myxobacteria and aerial hyphae formation by Streptomyces, or budding. Budding involves a cell forming a protrusion that breaks away and produces a daughter cell.\\
What are produced in cell division?\\
A: two identical clone daughter cells  The answer is two identical clone daughter cells.\\\\
\textcolor{darkgreen}{
Q: Janet’s ducks lay 16 eggs per day. She eats three for breakfast every morning and bakes muffins for her friends every day with four. She sells the remainder at the farmers' market daily for \$2 per fresh duck egg. How much in dollars does she make every day at the farmers' market?\\
A:}
\end{tcolorbox}
\end{figure*}

\begin{figure*}

\begin{tcolorbox}[
    title=Example Prompt for long-context GSM8k,
    width=\textwidth,
    colback=white,
    left=5pt,
    right=5pt,
    top=5pt,
    bottom=5pt
]
[...]
You are an expert Python programmer, and here is your task: Due to extreme variation in elevation, great variation occurs in the climatic conditions of Himachal . The climate varies from hot and subhumid tropical in the southern tracts to, with more elevation, cold, alpine, and glacial in the northern and eastern mountain ranges. The state has areas like Dharamsala that receive very heavy rainfall, as well as those like Lahaul and Spiti that are cold and almost rainless. 
Broadly, Himachal experiences three seasons: summer, winter, and rainy season. Summer lasts from mid-April till the end of June and most parts become very hot (except in the alpine zone which experiences a mild summer) with the average temperature ranging from 28 to 32 °C (82 to 90 °F). Winter lasts from late November till mid March. Snowfall is common in alpine tracts (generally above 2,200 metres (7,218 ft) i.e. in the higher and trans-Himalayan region).\\
What is the climate like? Your code should pass these tests:
\\
empty
\\
{[BEGIN]}\\
varies from hot and subhumid tropical\\
{[DONE]}
\\\\
\textcolor{darkgreen}{
You are an expert Python programmer, and here is your task: Write a function to find the similar elements from the given two tuple lists. Your code should pass these tests:\\
assert similar\_elements((3, 4, 5, 6),(5, 7, 4, 10)) == (4, 5)\\
assert similar\_elements((1, 2, 3, 4),(5, 4, 3, 7)) == (3, 4)\\
assert similar\_elements((11, 12, 14, 13),(17, 15, 14, 13)) == (13, 14)\\\\
{[BEGIN]}\\
def similar\_elements(test\_tup1, test\_tup2):\\
  res = tuple(set(test\_tup1) \& set(test\_tup2))\\
  return (res) \\
{[DONE]}}\\\\
You are an expert Python programmer, and here is your task: Unlike in multicellular organisms, increases in cell size (cell growth) and reproduction by cell division are tightly linked in unicellular organisms. Bacteria grow to a fixed size and then reproduce through binary fission, a form of asexual reproduction. Under optimal conditions, bacteria can grow and divide extremely rapidly, and bacterial populations can double as quickly as every 9.8 minutes. In cell division, two identical clone daughter cells are produced. Some bacteria, while still reproducing asexually, form more complex reproductive structures that help disperse the newly formed daughter cells. Examples include fruiting body formation by Myxobacteria and aerial hyphae formation by Streptomyces, or budding. Budding involves a cell forming a protrusion that breaks away and produces a daughter cell.\\
What are produced in cell division? Your code should pass these tests:\\
empty
\\
{[BEGIN]}\\
two identical clone daughter cells\\
{[DONE]}\\\\
\textcolor{darkgreen}{
You are an expert Python programmer, and here is your task: Write a python function to remove first and last occurrence of a given character from the string. Your code should pass these tests:\\
assert remove\_Occ("hello","l") == "heo"\\
assert remove\_Occ("abcda","a") == "bcd"\\
assert remove\_Occ("PHP","P") == "H"\\\\
{[BEGIN]}\\}

\end{tcolorbox}
\end{figure*}

\end{document}